\documentclass{article}

\usepackage[eandd,preprint]{neurips_2026}

\usepackage[utf8]{inputenc}
\usepackage[T1]{fontenc}
\usepackage{microtype}
\usepackage{graphicx}
\usepackage{subcaption}
\usepackage{booktabs}
\usepackage{multirow}
\usepackage{makecell}
\usepackage{array}
\usepackage{xcolor}
\usepackage{pifont}
\usepackage{tabularx}
\usepackage{seqsplit}
\usepackage{arydshln}
\usepackage{url}
\usepackage{hyperref}
\makeatletter
\renewcommand\hyper@natanchorstart[1]{}
\renewcommand\hyper@natanchorend{}
\renewcommand\hyper@natlinkstart[1]{}
\renewcommand\hyper@natlinkend{}
\renewcommand\hyper@natlinkbreak[2]{#1}
\makeatother
\usepackage{amsmath}
\usepackage{amsfonts}
\usepackage{amssymb}
\usepackage{mathtools}
\usepackage{amsthm}
\usepackage[capitalize,noabbrev]{cleveref}

\theoremstyle{plain}

\theoremstyle{definition}

\theoremstyle{remark}

\title{LongMemEval-V2: Evaluating Long-Term Agent Memory Toward Experienced Colleagues}

\author{%
Di Wu\thanks{Correspondence to \texttt{\{diwu,kwchang\}@cs.ucla.edu}.} \quad
Zixiang Ji \quad
Asmi Kawatkar \quad
Bryan Kwan \\[0.25em]
\textbf{Jia-Chen Gu} \quad
\textbf{Nanyun Peng} \quad
\textbf{Kai-Wei Chang} \\[0.25em]
University of California, Los Angeles \\[0.25em]
\url{https://xiaowu0162.github.io/longmemeval-v2/}
}
\begin{document}

\maketitle

\begin{abstract}

Long-term memory is crucial for agents in specialized web environments, where success depends on recalling interface affordances, state dynamics, workflows, and recurring failure modes. However, existing memory benchmarks for agents mostly focus on user histories, short traces, or downstream task success, leaving open how to directly evaluate whether memory systems effectively internalize environment-specific experience. To address this gap, we introduce LongMemEval-V2 (LME-V2), a benchmark for evaluating whether memory systems can help agents acquire the experience needed to become knowledgeable colleagues in customized environments. LME-V2 contains 451 manually curated questions covering five core memory abilities for web agents: \textit{static state recall}, \textit{dynamic state tracking}, \textit{workflow knowledge}, \textit{environment gotchas}, and \textit{premise awareness}. These questions are paired with history trajectories containing up to 500 trajectories and 115M tokens. We use a context gathering formulation: memory systems consume history trajectories and return compact evidence for downstream question answering. As initial baselines for this challenging setting, we propose a suite of two memory methods: AgentRunbook-R, an efficient RAG-based memory with knowledge pools for raw state observations, events, and strategy notes, and AgentRunbook-C, which stores trajectories as files and invokes a coding agent to gather evidence in an augmented sandbox. Experiments show that AgentRunbook-C achieves the best performance with 72.5\% average accuracy, outperforming the strongest RAG baseline (48.5\%) and the off-the-shelf coding agent baseline (69.3\%). Despite the strong performance gains, coding agent based methods have high latency costs. While AgentRunbook-C advances the accuracy-latency Pareto frontier, substantial room for improvement remains. Together, these results establish LME-V2 as a challenging testbed for developing long-term memory systems that turn accumulated agent trajectories into reusable environment experience.

\end{abstract}

\section{Introduction}

Long-term memory helps large language models (LLMs) operate beyond their context and parameters by storing and recalling information over long horizons \citep{memorizing_transformers,memgpt,memoryllm}. Memory is especially important for agent systems, where LLMs interact with specialized environments over many steps. Recent works show that memorizing task procedures, interface affordances, and hidden failure modes improve agent performance at inference time \citep{voyager, expel, repairagent, awm, chemagent}.

However, benchmarks for memory in the agentic context remain limited. Existing memory works mainly evaluate retrieval and reasoning over long documents or user chat histories \citep{RULER, LongBenchV2, longmemeval, LoCoMo, BEAM}. Recent works consider evaluating memorization over agent trajectories, but often use simplified game environments \citep{AgentLongBench, li2026emembench}, emphasize limited dependencies within one or a few trajectories \citep{he2026memoryarena, zhao2026amabench}, or evaluate indirectly through downstream task success \citep{he2026memoryarena}. As a result, they provide limited insight into whether memory systems can accumulate holistic, environment-specific knowledge from sustained interaction with a complex environment. To highlight this perspective, this paper uses the following framing:
\begin{center}
    \textit{A high-quality memory makes an agent an experienced colleague in a specialized environment.}
\end{center}

\begin{figure*}[t!]
    \centering
    \includegraphics[width=\linewidth]{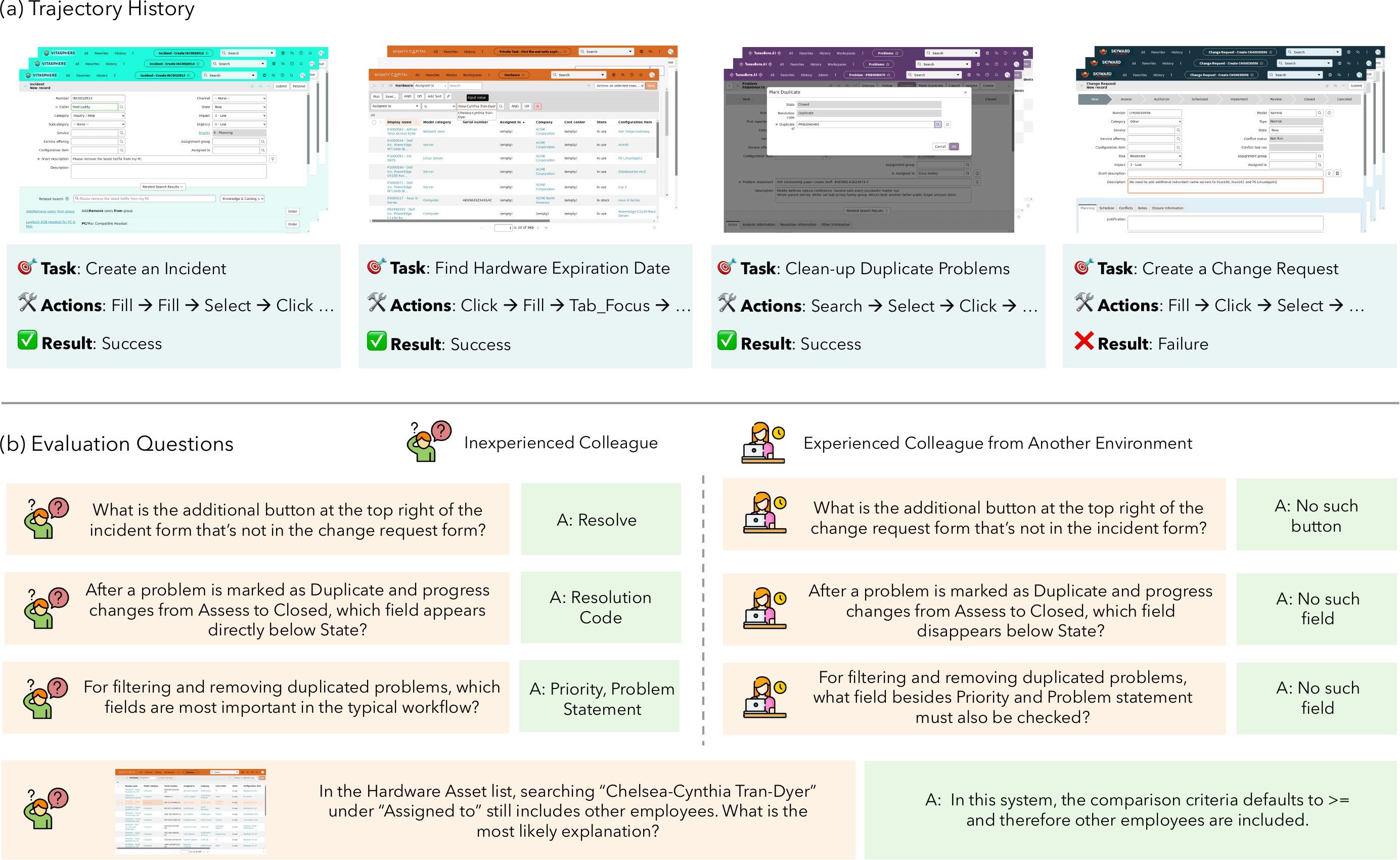}
    \caption{Examples of LongMemEval-V2 questions. We display examples from the WorkArena domain. LME-V2 questions exercise diverse memory abilities: static (row 1), dynamic (row 2), workflow (row 3), gotchas (row 4), and premise awareness (right column).}
    \label{fig:main-examples}
\end{figure*}

Driven by this view, we introduce LongMemEval-V2 (LME-V2), a benchmark for evaluating whether memory systems can help web agents acquire the experience needed to become knowledgeable colleagues. LME-V2 leverages customized websites including Magento shopping, shopping admin, Postmill forum, and ServiceNow from WebArena \citep{webarena} and WorkArena \citep{workarena, workarena++}. From task-solving web agent trajectories, we manually curate 451 questions covering five core memory abilities: static state recall, dynamic state tracking, workflow knowledge, environment gotchas, and premise awareness. We provide examples in \Cref{fig:main-examples} and ability definitions in \S\ref{section-ability-definition}. These questions are specific to the customized environments and thus remain generally unanswerable by recent frontier LLMs (\S\ref{section-pilot-studies}). LME-V2 further pairs the questions with a sequence of web agent trajectories (``haystacks'', following \citet{kamradt2023needle}), where only a small fraction bears the answers to each question (``needles''). LME-V2-Small provides a 100-trajectory haystack shared by all questions, and LME-V2-Medium has 500-trajectory question-specific haystacks. Compared to prior benchmarks, LME-V2 poses new challenges with its deep context (25M/115M tokens in the small/medium tiers) and comprehensive memory ability coverage (\Cref{tab:benchmark-comparison}). 

LME-V2 evaluates memory with a \textit{context gathering} formulation (\S\ref{section-evalation-formuation}). A memory system implements two APIs: \texttt{Insert}, which consumes a trajectory, and \texttt{Query}, which returns a multimodal memory context for a question. For each question, we stream the associated trajectories sequentially into memory, invoke \texttt{Query}, truncate the returned context to a fixed token budget, and ask a fixed reader LLM to answer. This provides a direct evaluation of memory quality with a practical interface that a downstream agent would use. We report both answer accuracy and query latency.

\newcommand{\yes}{\ding{51}}
\newcommand{\no}{\ding{55}}
\newcommand{\unk}{?}
\newcommand{\na}{N/A}

\begin{table*}[ht!]
\centering
\small
\setlength{\tabcolsep}{2pt}
\renewcommand{\arraystretch}{1.12}
\caption{Comparison with existing memory and long-context benchmarks. \# Sess., \# Tok., and \# Q denote the history size in sessions or tokens, as well as the total number of questions. MM denotes whether the context or question is a multimodal mixture of text and images. For benchmarks with multiple preset length tiers, we report a range between the minimum length tier and the maximum length tier. For the other benchmarks, we report the averaged context size over all examples.}
\resizebox{\linewidth}{!}{
\begin{tabular}{@{}lccccccccccc@{}}
\toprule
\multirow{2}{*}{\textbf{Benchmark}} &
\multirow{2}{*}{\textbf{Domain}} &
\multicolumn{3}{c}{\textbf{Context Profile}} &
\multicolumn{2}{c}{\textbf{Question Profile}} &
\multicolumn{5}{c}{\textbf{Memory Ability}} \\
\cmidrule(lr){3-5}
\cmidrule(lr){6-7}
\cmidrule(l){8-12}
& &
\makecell{\# Sess.} &
\makecell{\# Tok.} &
\makecell{MM} &
\makecell{\# Q} &
\makecell{MM} &
\makecell{Static} &
\makecell{Dynamic} &
\makecell{Workflows} &
\makecell{Gotchas} &
\makecell{Premise} \\
\midrule

\multicolumn{12}{@{}l}{\textit{General Long Context}} \\
\addlinespace[1pt]
LongBench V2
& Mixed & \na & 260k & \no & 503 & \no & \yes & \no & \no & \no & \no \\
MemoryAgentBench
& Mixed & \na & 285k & \no & 2,071 & \no & \yes & \no & \no & \no & \no \\
CL-Bench
& Mixed & \na & 10k & \no & 1,899 & \no & \yes & \no & \yes & \no & \no \\

\midrule
\multicolumn{12}{@{}l}{\textit{Conversational Long Context}} \\
\addlinespace[1pt]
LoCoMo
& User-user chat & 28 & $\sim$16k & \yes & 7,512 & \no & \yes & \yes & \no & \no & \yes \\
LongMemEval-V1
& User-assistant chat & 48--475 & 115k--1.5M & \no & 500 & \no & \yes & \yes & \no & \no & \yes \\
PersonaMem
& User-assistant chat & 5--60 & 26k--951k & \no & 5,990 & \no & \yes & \yes & \no & \no & \no \\
PersonaMem-v2
& User-assistant chat & 10--20 & 33k--124k & \yes & 5,000 & \no & \yes & \yes & \no & \no & \no \\
BEAM
& User-assistant chat & 4.5--100 & 124k--10M & \no & 2,000 & \no & \yes & \yes & \no & \no & \yes \\

\midrule
\multicolumn{12}{@{}l}{\textit{Agentic Long Context}} \\
\addlinespace[1pt]
MemoryArena
& Agent (mixed) & 7 & 40k+ & \no & 766 & \no & \yes & \no & \yes & \yes & \no \\
AgentLongBench
& Game agent & 1 & 31k--4M & \no & 6,400 & \no & \yes & \yes & \no & \no & \no \\
EMemBench
& Game agent & 1 & 2k--$\infty$ & \yes & 1,280+ & \no & \yes & \no & \yes & \yes & \yes \\
FileGramBench & File-system agent & 12 & 11k & \yes & 4,333 & \no & \yes & \no & \yes & \no & \no \\
AMA-Bench
& Agent (mixed) & 1 & 57k & \no & 2,496 & \no & \yes & \yes & \yes & \yes & \no \\
\textbf{LongMemEval-V2}
& \textbf{Web agent} & \textbf{100--498} & \textbf{25M--115M} & \yes & \textbf{451} & \yes & \yes & \yes & \yes & \yes & \yes \\

\bottomrule
\end{tabular}
}
\label{tab:benchmark-comparison}
\end{table*}

To succeed in LME-V2, a memory system needs to intelligently store and filter information from the noisy agent trajectories, retaining both low-level observations as well as higher-level environment dynamics and procedural knowledge. As a result, naive application of popular agent memory methods could be ineffective as they are biased towards less noisy conversational contexts \citep{mem0} or high-level strategic knowledge \citep{awm, ouyang2025reasoningbank}. In this paper, we propose AgentRunbook, a simple yet effective baseline consisting of two variants, optimized separately for efficiency and accuracy. AgentRunbook-R is an efficient retrieval-augmented generation (RAG) pipeline inspired by agentic memory works such as \citet{amem}. It prompts an LLM controller to update and to actively query three knowledge pools: raw observations, state transition events, and high-level strategy notes
(\S\ref{section-approach-RAG}). 
AgentRunbook-R is efficient and covers major memory abilities, but its simple design is not optimized for detailed evidence selection. Inspired by \citet{coding-agent-long-context}, we propose AgentRunbook-C, a coding agent-based memory method that casts memory management as a file management problem. AgentRunbook-C stores raw trajectories directly as files. At query time, it augments an off-the-shelf coding agent harness with workflow documents, memory manifests, and helper scripts, then invokes the agent to assemble a compact evidence set (\S\ref{section-approach-CA}).

We evaluate the memory designs on the small and medium tiers of LME-V2. To begin with, a simple RAG method that retrieves state slices can only achieve an overall acccuracy of 40.1\%, and AgentRunbook-R further improves to 57.8\%. Accuracy-wise, we find the off-the-shelf Codex agent \citep{openaiCodexCLI} has competitive performance, achieving a surprisingly high 69.3\% accuracy. However, the agent achieves this at a cost of about 182 seconds per query, about 6.9 times slower than AgentRunbook-R. With our specialization designs, AgentRunbook-C performs best overall with 72.5\% accuracy while being 32\% faster than Codex at query time. Our further analyses reveal that AgentRunbook-C significantly advances the accuracy-latency frontier, but the room for future improvement remains large (\S\ref{results-accuracy-latency}). Overall, LME-V2 formulates a new standard for agent memory evaluation and provides a concrete testbed for memory modules that make long-running agents more reliable, adaptive, and useful in real-world environments.

\section{Related Work}

\paragraph{Long-Context and Memory Evaluation}

Long-term memory evaluation can be seen as part of the long-lasting effort to evaluate LLMs and retrieval systems on recalling information across extended context. Early line of benchmarks focuses on testing information retrieval, aggregation, and instruction following over long input documents \citep{RULER, NOCHA, NoLiMa, LongBenchV2, CL-bench}. Subsequent work expanded the focus to personalized memory, covering explicit user facts and implicit preferences, with benchmarks such as LoCoMo \citep{LoCoMo}, DialSim \citep{DialSim}, PerLTQA \citep{PerLTQA}, LongMemEval \citep{longmemeval}, PersonaMem \citep{personamem-v1, personamem-v2}, and BEAM \citep{BEAM}. LMEB \citep{LMEB} isolates the retrieval component and evaluates dense retrievers on memory workloads. In contrast, among a new series of efforts, LME-V2 targets experience memory with context constructed from web agent history trajectories. This shift introduces substantially more complex contexts, a new ability taxonomy, and memory designs centered on agent experience.

\paragraph{Memory Systems for Agents}
As LLM agents tackle long-horizon tasks in complex environments, memory becomes important both for recalling earlier detailed trajectory context \citep{hu2024hiagent} and for consolidating high-level knowledge across trajectories \citep{awm, ouyang2025reasoningbank, wu2025autoscalingcontinuousmemory}. Memory has also been linked to improving inference-time performance through extended exploration and sleep-time offline consolidation \citep{ouyang2025reasoningbank, lin2025sleeptimecompute}. Despite this progress, direct evaluation of memory quality in agent settings remains limited. MemoryArena \citep{he2026memoryarena} measures memory indirectly through the success rate of interdependent task sequences. AgentLongBench \citep{AgentLongBench} and EMemBench \citep{li2026emembench} use synthetic agent histories and test recall of details from those traces. FileGram \citep{filegram} studies reasoning over file system behavior traces. AMA-Bench \citep{zhao2026amabench} is closest to our setting, as it curates questions from agent trajectories in diverse domains such as embodied, web, and gaming agents. However, AMA-Bench focuses on understanding one trajectory, while LME-V2 focuses on environment knowledge induced across many past trajectories. To our knowledge, LME-V2 is also the first benchmark in this setting to scale the history length to tens or even over 100 million tokens.

\paragraph{Agents as Memory Controllers}
Recent work on agentic memory proposes memory systems in which memory write and read operations are controlled by an LLM rather than a fixed pipeline. MemGPT \citep{memgpt} and StateLM \citep{statelm} enable models to manage context programmatically. A-MEM \citep{amem} and Mem0 \citep{mem0} introduce scaffolding that allows an LLM to evolve memory content and structure over time. Memory-R1 \citep{memory-r1} and Mem-$\alpha$ \citep{memory-alpha} learn memory update actions via reinforcement learning. MemSkill \citep{memskill} learns memory skills to guide memory update behavior at a finer granularity. In this work, we further expand the notion of agentic memory. Inspired by \citet{coding-agent-long-context} and \citet{cocoabench}, we view a general coding agent with tool use and file system manipulation abilities as a strong controller for file-based memory. Based on this perspective, we design AgentRunbook-C, which augments an off-the-shelf coding agent harness with workflow documents, query-time rendered artifacts, and helper scripts, yielding a strong accuracy-latency trade-off on LME-V2.

\section{LongMemEval-V2}

\subsection{Core Memory Ability Definition}
\label{section-ability-definition}

What does an experienced colleague internalize after repeatedly working in an environment? We categorize the learned experience into five memory abilities:

\begin{itemize}
    \item \textbf{Static State Recall}. An experienced colleague remembers important landmarks, page layouts, module affordances, and subtle differences across states.
    \item \textbf{Dynamic State Tracking}. An experienced colleague can act as a world model of the environment: given states and actions, they understand how the environment changes.
    \item \textbf{Workflow Knowledge}. An experienced colleague knows the steps needed to perform common tasks in the customized environment.
    \item \textbf{Environment Gotchas}. An experienced colleague is aware of common recurring issues in the current environment and can avoid environment-specific failures.
    \item \textbf{Premise Awareness}. An experienced colleague can recognize assumptions that are valid in another environment but wrong in the current one.
\end{itemize}

\subsection{Annotation}

To holistically evaluate these memory abilities, we curate LongMemEval-V2 from multimodal web agent trajectories. The annotation has four steps: trajectory collection, question annotation, answer trajectory labeling, and haystack creation. We present full details in Appendix \cref{appendix-further-benchmark-details}.

\paragraph{Trajectory Collection} We collect trajectories from three web agent benchmarks: WebArena \citep{webarena}, WorkArena \citep{workarena}, and WorkArena++ \citep{workarena++}, leveraging their  \textbf{OneStopShop}, \textbf{CMS}, \textbf{Reddit}, \textbf{ServiceNow} environments. The trajectories are collected using the AgentLab\footnote{\url{https://github.com/ServiceNow/AgentLab}.} library, which provides unified state representations, action spaces, and a ReAct-style base agent implementation \citep{react}. Using the base agent and Codex \citep{openaiCodexCLI}, we perform rejection sampling with GPT-5.2 \citep{openai2025gpt52} and GPT-5-mini \citep{openai2026gpt5} as the LLMs. The final pool contains 599 trajectories from WebArena and 941 from WorkArena/WorkArena++. The overall success rate is 52.0\%, and each trajectory contains 28.1 states on average. 

\begin{figure*}[t!]
    \centering
    \includegraphics[width=0.95\linewidth]{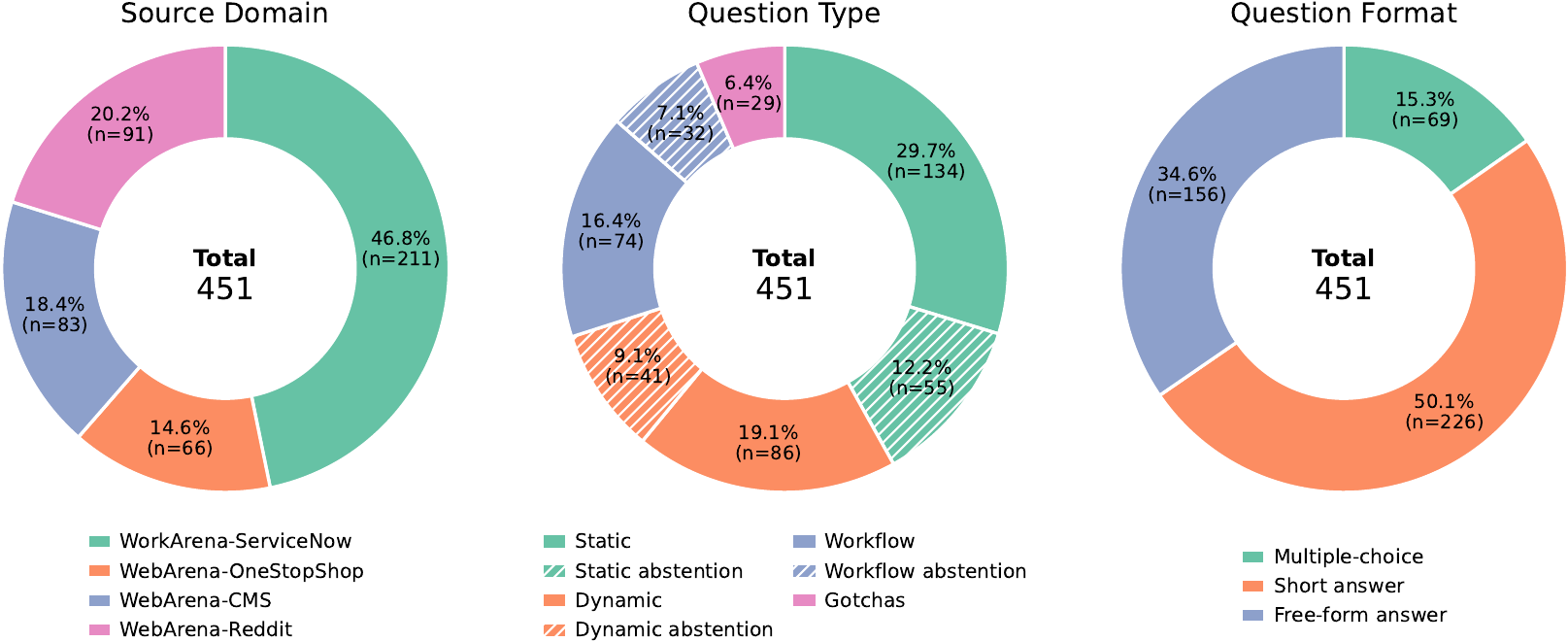}
    \caption{LME-V2 questions cover diverse source domains, question types, and formats.}
    \label{fig:question-distribution}
\end{figure*}

\paragraph{Question Annotation} All questions are constructed through \textbf{manual annotation}. Following the memory ability taxonomy, human experts first inspect the trajectories to identify  various information an experienced colleague would naturally learn. We then curate and filter questions to ensure strong proprietary LLMs cannot answer from parametric knowledge alone\footnote{We manually tested Gemini-3-Pro \citep{google2025gemini3pro}, GPT-5.2 \citep{openai2025gpt52}, Grok-4.1-thinking \citep{xai2025grok41}, and Claude-Opus-4.6 \citep{anthropic2026claudeopus46} and ensured that at least two out of four models answered the questions incorrectly.}. Gotchas questions are framed as scenarios where an inexperienced worker sends a message with a screenshot, while the other questions are expressed as text-only true/false, multiple choice, or short answer questions. Finally, based on existing static, dynamic, and workflow questions, we curate abstention questions with wrong premises that the model must identify to succeed. \Cref{fig:main-examples} shows example questions in each category. \Cref{fig:question-distribution} presents source domain, type, and format distribution of the final question pool. On average, questions require 1.4 trajectories to answer (min 1, max 5). However, many dynamic and workflow questions require evidence synthesized from many states within a supporting trajectory.

\paragraph{Answer Trajectory Labeling} During annotation, annotators identify a seed set of answer-bearing trajectories for each question. To construct shared history haystacks where we can jointly minimize the number of answer-bearing trajectories for all questions, we perform additional annotation to label all trajectories that contain the answer for each question. We use the Codex coding agent to generate initial proposals. Human experts then verify that the question-trajectory correspondence for trajectories included in the final core haystack set. We provide details in \Cref{appendix-further-benchmark-details}.

\begin{figure*}[t!]
    \centering
    \includegraphics[width=0.99\linewidth]{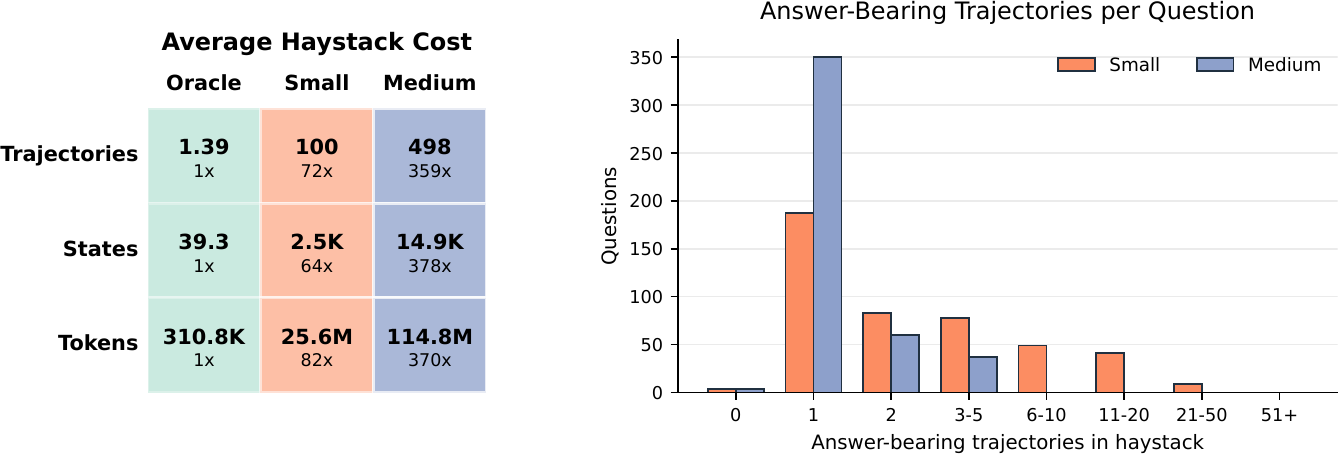}
    \caption{Basic statistics of history haystacks in LME-V2. The small and medium tiers cover a large number of states and tokens (left). Meanwhile, the trajectories containing answers to each question remain generally sparse in the haystack (right).}
    \label{fig:haystack-trajectory-distribution}
\end{figure*}

\paragraph{Haystack Creation} Based on the answer trajectory labels, we programmatically assemble two tiers of history trajectory haystacks: a small variant that contains 100 trajectories shared by all questions, and a medium variant that contains roughly 500 trajectories per question. We refer to them as LME-V2-Small and LME-V2-Medium for the rest of the paper. For LME-V2-Small, we create one haystack for the ServiceNow questions and one haystack for the WebArena domains. All haystacks contain a balanced ratio of successful and failed trajectories, and many questions can only be answered from failed trajectories. \Cref{fig:haystack-trajectory-distribution} presents further statistics of the haystacks. The final history lengths of LME-V2-Small and LME-V2-Medium are approximately 25M and 115M tokens, while each question's answer-bearing trajectory set remains sparse in the haystack.

\Cref{tab:benchmark-comparison} compares LME-V2 with previous long-term memory benchmarks. LME-V2 has substantially longer histories than prior long-term memory benchmarks, naturally includes multimodal evaluation, and provides a broad coverage of crucial agent memory capabilities.

\subsection{Evaluation Formulation}
\label{section-evalation-formuation}

We formulate LME-V2 as a \textbf{context gathering} task. For each question $q_i$ with gold answer $y_i$, a memory system receives an ordered trajectory haystack $\mathcal{H}_i=\{h_{i,1},\ldots,h_{i,m_i}\}$, where each $h_{i,j}$ is a trajectory.
The system must support two APIs, $\texttt{Insert}(h)$ and $\texttt{Query}(q)$.
We sequentially insert all trajectories in $\mathcal{H}_i$, query the final memory with $q_i$, and obtain a returned context $c_i$:
\[
\begin{aligned}
\mathcal{M}_{i,j} &= \texttt{Insert}_{\mathcal{M}_{i,j-1}}(h_{i,j}), \\
c_i &= \texttt{Query}_{\mathcal{M}_{i,m_i}}(q_i).
\end{aligned}
\]
A fixed reader model $R$ answers from the question and a bounded memory context: $\hat{y}_i = R(q_i, \mathrm{Trunc}(c_i))$\footnote{We set the truncation budget to 200k tokens empirically.}. We report answer accuracy and query latency. Accuracy is computed by normalized string matching for structured answers and an LLM judge for free-form answers.

\subsection{Pilot Studies}
\label{section-pilot-studies}

\begin{figure*}[t]
    \centering
    \begin{minipage}[t]{0.35\textwidth}
        \vspace{0pt}
        \centering
        \includegraphics[width=\linewidth]{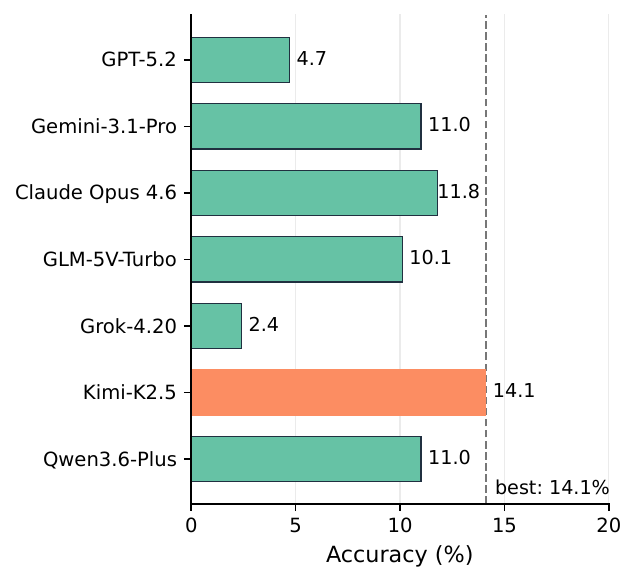}

    \end{minipage}
    \hfill
    \begin{minipage}[t]{0.62\textwidth}
        \vspace{0pt}
        \centering
        \small
        \setlength{\tabcolsep}{3.0pt}
        \renewcommand{\arraystretch}{1.03}

        \resizebox{\linewidth}{!}{%
        \begin{tabular}{@{}lccccc@{}}
        \toprule
        \textbf{Context} & \textbf{Overall} & \textbf{Static} & \textbf{Dynamic} & \textbf{Workflow} & \textbf{Gotchas} \\
        \midrule
        \multicolumn{6}{@{}l}{\textit{\textbf{Qwen3.5-9B (thinking enabled)}}} \\
        \quad No context & 0.016 & 0.000 & 0.015 & 0.155 & 0.136 \\
        \quad Oracle trajectories & 0.596 & 0.566 & 0.668 & 0.718 & 0.310 \\
        \quad Oracle slices + notes & 0.825 & 0.908 & 0.879 & 0.750 & 0.484 \\
        \midrule
        \multicolumn{6}{@{}l}{\textit{\textbf{GPT-5.4-mini (medium reasoning)}}} \\
        \quad No context & 0.045 & 0.025 & 0.010 & 0.075 & 0.171 \\
        \quad Oracle trajectories & 0.653 & 0.660 & 0.696 & 0.697 & 0.484 \\
        \quad Oracle slices + notes & 0.863 & 0.950 & 0.843 & \textbf{0.905} & 0.467 \\
        \midrule
        \multicolumn{6}{@{}l}{\textit{\textbf{Codex + GPT-5.4-mini (xhigh reasoning)}}} \\
        \quad Oracle trajectory files & \textbf{0.897} & \textbf{0.986} & \textbf{0.947} & 0.815 & \textbf{0.517} \\
        \bottomrule
        \end{tabular}
        \vspace{2mm}
        }
        
    \end{minipage}

    \caption{
    Pilot studies on LME-V2 non-abstention questions.
    (a) Frontier LLMs perform poorly without trajectory history, suggesting that parametric knowledge alone is insufficient for LME-V2.
    (b) LME-V2 is challenging to answer even with oracle answer-bearing trajectories and optimizations such as evidence slicing with notes or using a coding agent harness help improve performance.
    }
    \label{fig:pilot_studies}
\end{figure*}

We perform two pilot studies. First, we  evaluate whether LME-V2 questions require environment-specific trajectory evidence. Then, we sanity check whether answer-bearing trajectories are sufficient for reliable question answering. These studies use a direct question answering setup rather than the context gathering formulation used in the main experiments, and evaluate non-abstention questions only. Full per-category results, prompts, and sandbox instructions are provided in Appendix \Cref{appendix-pilot-study}.

To begin with, can recent frontier LLMs answer LME-V2 questions without the trajectory history? We prompt strong LLMs with only the question. As shown in \Cref{fig:pilot_studies} (left), all LLMs perform poorly in this setting: the best model reaches only 14.1\% overall accuracy, suggesting that LME-V2 questions generally cannot be answered from public or parametric knowledge alone.

Second, we give models oracle access to the answer-bearing trajectories to isolate the difficulty of reading and grounding trajectory evidence. Long-context prompting shows much higher accuracy but remains limited due to the trajectory size exceeding the model's context window. We further consider two techniques: 1) annotating ground-truth states containing the evidence and providing only radius-1 evidence slices around them and 2) summarizing strategy notes containing important procedures and gotchas identified in the trajectory. These two techniques further improve direct QA to 82.5\% and 86.3\%, respectively. Finally, we represent the trajectories as files and use the off-the-shelf Codex coding agent to directly answer the question. Surprisingly, GPT-5.4-mini with the Codex harness answers the questions better than prompting approach, suggesting that detailed evidence inspection via multi-step tool use is effective for understanding agent trajectories, and that coding agents might have good performance acting as memory controllers. Overall, these findings confirm that the answer trajectory labelings are accurate enough and motivate our memory method design.

\section{AgentRunbook}
\label{section-approach}

\begin{figure*}[t!]
    \centering
    \includegraphics[width=0.99\linewidth]{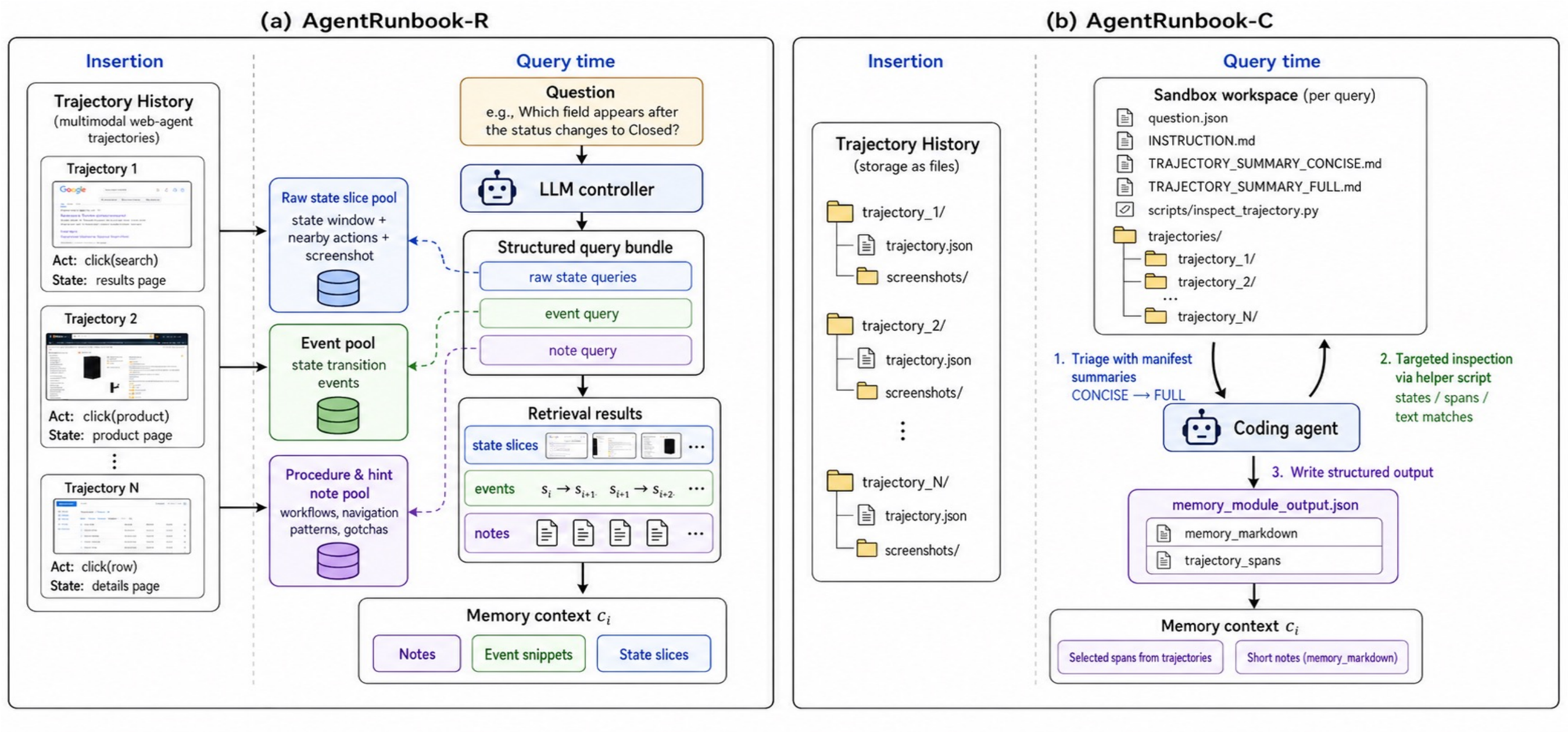}
    \caption{An illustration of the AgentRunbook memory modules. (a) AgentRunbook-R digests trajectories at insertion time and creates dedicated knowledge pools for raw states, events, and notes. At query time, an LLM controller generates multi-stream queries for each pool. (b) AgentRunbook-C stores trajectories as files at insertion time and constructs a sandbox workspace with specialized instructions and manifests for a coding agent to collect the required evidence efficiently.}
    \label{fig:agentrunbook-methods}
\end{figure*}

LME-V2 is challenging because the evidence needed for a question can mix low-level UI observations, state transitions, and reusable task procedures. Memory modules therefore need to organize noisy agent trajectories into compact representations and index them for targeted recall. We propose two memory designs: AgentRunbook-R, a structured RAG pipeline with separate knowledge pools, and AgentRunbook-C, a coding agent based method that casts memorizing agentic contexts as a file management problem. \Cref{fig:agentrunbook-methods} illustrates the workflow of both methods. 

\subsection{AgentRunbook-R}
\label{section-approach-RAG}
AgentRunbook-R, where R denotes RAG, extracts structured memory items at insertion time and retrieves them at query time. To recall information at different granularities, AgentRunbook-R uses separate knowledge pools and a retrieval mechanism over these pools. Given a trajectory $h$, AgentRunbook-R builds three memory pools. The \textbf{raw state slice} pool stores windows centered at trajectory states, including local UI observations and nearby actions. This pool preserves fine-grained visual and textual evidence. The \textbf{state transition event} pool stores events extracted from consecutive states. These events describe how actions change the environment, accumulating evidence for an environment world model. The \textbf{procedure and hint note} pool stores trajectory-level notes that capture reusable workflows, navigation patterns, and environment-specific gotchas. This pool is inspired by prior works that consolidate trajectory experience into compact reusable knowledge \citep{awm,ouyang2025reasoningbank}.

At query time, AgentRunbook-R uses an LLM controller to reason about the query and the current memory snapshot, then generate retrieval queries for the knowledge pools: multiple raw state queries for exact UI evidence, one event query for important state changes, and one note query for procedural knowledge. The controller may skip irrelevant streams. Each query retrieves from its corresponding pool using dense retrieval, and the results are rendered as a multimodal memory context $c_i$. This design keeps query latency low while supporting evidence at different granularities.

\subsection{AgentRunbook-C}
\label{section-approach-CA}

AgentRunbook-C, where C denotes coding agent, is motivated by the observation that general coding agents are effective file system manipulators and tool users \citep{coding-agent-long-context}. Rather than compressing retrieval behavior into a fixed vector search pipeline, AgentRunbook-C stores trajectories directly as files and uses a coding agent to search, inspect, and select evidence at query time.

An off-the-shelf coding agent, however, is not optimized as a memory module. It may over-explore, under-explore, or inspect the data inefficiently. AgentRunbook-C adds three lightweight scaffolding components to the coding agent. First, a \textbf{workflow document} instructs the agent to act as a memory module and outlines the steps for collecting evidence. Second, query-time \textbf{manifest artifacts} summarize the current memory layout, helping the agent shortlist relevant trajectories before detailed inspection. Third, a \textbf{helper script} exposes common trajectory inspection operations, such as viewing a state span or searching within a trajectory. At insertion time, AgentRunbook-C stores each trajectory on disk. At query time, it creates a sandbox with the question, workflow document, helper script, and rendered manifest artifacts. The coding agent then writes a structured retrieval output containing a short memory note and selected trajectory state spans, which are rendered into $c_i$.

\section{Experiments}

We evaluate all methods on LME-V2 under the context gathering formulation. The returned memory context is truncated to 200K tokens and answered by a fixed Qwen3.5-9B reader \citep{qwen35}. For RAG methods, we use Qwen3.5-9B as the memory controller and Qwen3-Embedding-8B for retrieval. For coding agent methods, we use Codex and GPT-5.4-mini with different reasoning efforts. The full implementation details of AgentRunbook and baselines are provided in \Cref{app:agentrunbook_details}.

\subsection{Main Results}
\label{results-main}

\newcommand{\dingup}[1]{\textsuperscript{\ding{#1}}}

\begin{table*}[t]
\centering
\small
\setlength{\tabcolsep}{4pt}
\renewcommand{\arraystretch}{1.08}
\caption{Main results with baselines and ablations. The downstream reader is always Qwen3.5-9B. We boldface the best results in each method family. \dingup{67} means statistically significantly outperforming the non-ablation baselines via paired bootstrap test ($p<0.05$). AgentRunbook strongly outperforms the baseline in RAG family and achieves a superior latency in coding agent family. }
\label{tab:main_results}
\resizebox{\linewidth}{!}{
\begin{tabular}{lcccccccccccc}
\toprule
\multirow{2}{*}{\textbf{Method}} &
\multicolumn{6}{c}{\textbf{LME-V2-Small}} &
\multicolumn{6}{c}{\textbf{LME-V2-Medium}} \\
\cmidrule(lr){2-7}
\cmidrule(lr){8-13}
& \textbf{Overall} & \textbf{Static} & \textbf{Dynamic} & \textbf{Workflow} & \textbf{Gotchas} & \textbf{Latency}
& \textbf{Overall} & \textbf{Static} & \textbf{Dynamic} & \textbf{Workflow} & \textbf{Gotchas} & \textbf{Latency} \\
\midrule
No retrieval
& 0.013 & 0.000 & 0.008 & 0.094 & 0.138 & 0s
& 0.013 & 0.000 & 0.008 & 0.094 & 0.138 & 0s \\
\midrule
\multicolumn{13}{l}{\textit{\textbf{RAG Methods (Controller = Qwen3.5 9B, thinking enabled)}}} \\
RAG: query $\rightarrow$ slice
& 0.428 & 0.471 & 0.425 & 0.415 & 0.207 & 0.1s
& 0.381 & 0.434 & 0.405 & 0.293 & 0.242 & 0.1s \\
RAG: query $\rightarrow$ slice + notes
& 0.510 & 0.524 & 0.496 & 0.528 & 0.414 & 0.2s
& 0.459 & 0.487 & 0.472 & 0.434 & 0.310 & 0.3s \\
AgentRunbook-R
& \textbf{0.586}\dingup{67} & \textbf{0.661}\dingup{67} & 0.583\dingup{67} & 0.528 & 0.310 & 26.9s
& \textbf{0.570}\dingup{67} & \textbf{0.630}\dingup{67} & \textbf{0.614}\dingup{67} & \textbf{0.472}\dingup{67} & \textbf{0.345} & 25.8s \\
\quad\quad\quad\quad\quad\quad -- raw slice pool
& 0.423 & 0.286 & 0.551 & \textbf{0.538} & \textbf{0.345} & 16.7s
& 0.335 & 0.233 & 0.433 & 0.377 & 0.413 & 17.1s \\
\quad\quad\quad\quad\quad\quad -- event pool
& 0.556 & 0.614 & 0.559 & 0.528 & 0.276 & 19.1s
& 0.484 & 0.534 & 0.496 & 0.434 & 0.276 & 18.5s \\
\quad\quad\quad\quad\quad\quad -- note pool
& 0.579 & 0.651 & \textbf{0.614} & 0.481 & 0.310 & 22.8s
& 0.499 & 0.561 & 0.543 & 0.396 & 0.276 & 20.5s \\
\midrule
\multicolumn{13}{l}{\textit{\textbf{Coding Agent Methods (Controller = GPT-5.4-mini, xhigh reasoning)}}} \\
Codex
& 0.699 & 0.804 & 0.670 & 0.575 & \textbf{0.586} & 177.2s
& 0.687 & 0.783 & 0.646 & 0.613 & \textbf{0.517} & 185.8s \\
AgentRunbook-C
& \textbf{0.749}\dingup{67} & 0.820\dingup{67} & \textbf{0.724}\dingup{67} & \textbf{0.726}\dingup{67} & 0.483 & 108.3s
& 0.701 & 0.788 & \textbf{0.701}\dingup{67} & 0.613 & 0.449 & 139.9s \\
\quad\quad\quad\quad\quad\quad -- workflow
& 0.701 & 0.772 & 0.677 & 0.632 & \textbf{0.586} & 167.9s
& 0.641 & 0.709 & 0.646 & 0.575 & 0.414 & 231.9s \\
\quad\quad\quad\quad\quad\quad -- manifest artifacts
& 0.747 & \textbf{0.847} & 0.709 & 0.698 & 0.448 & 155.0s
& 0.681 & 0.767 & 0.685 & 0.576 & 0.483 & 211.6s \\
\quad\quad\quad\quad\quad\quad -- helper functions
& 0.714 & 0.783 & \textbf{0.724} & 0.660 & 0.414 & 145.9s
& \textbf{0.718} & \textbf{0.804} & 0.693 & \textbf{0.689} & 0.380 & 182.5s \\
\bottomrule
\end{tabular}
}
\end{table*}

As shown in \Cref{tab:main_results}, the no-retrieval baseline is near zero, confirming that the reader cannot answer without memory context. A simple query-to-slice RAG baseline reaches 42.8\% on LME-V2-Small and 38.1\% on LME-V2-Medium, while adding trajectory notes improves performance to 51.0\% and 45.9\%. AgentRunbook-R further improves over the strongest RAG baseline, reaching 58.6\% on LME-V2-Small and 57.0\% on LME-V2-Medium. The ablations show that the raw slice pool is important for static questions, while removing the event pool harms static, dynamic, and gotchas questions. Workflow questions benefit from consolidating trajectory experience into reusable events and notes rather than only retrieving local observations.

AgentRunbook-C achieves the best overall accuracy, reaching 74.9\% on LME-V2-Small and 70.1\% on LME-V2-Medium. It also improves over vanilla Codex, which reaches 69.9\% and 68.7\%, respectively. The ablations show that workflow instructions are consistently important, while manifest artifacts mainly improve efficiency. Helper functions affect the performance in a mixed way: they improve the small-tier result and reduce latency compared with the most expensive ablations, but their effect on medium-tier accuracy is not uniformly positive. Overall, the results show that coding agents can serve as strong memory controllers with a proper file-based environment. In \Cref{appendix-further-analysis}, we further analyze the error patterns and the tool calling behavior of AgentRunbook-C.

\subsection{Accuracy and Latency Trade-off}
\label{results-accuracy-latency}

\begin{figure*}[t!]
    \centering
    \includegraphics[width=\linewidth]{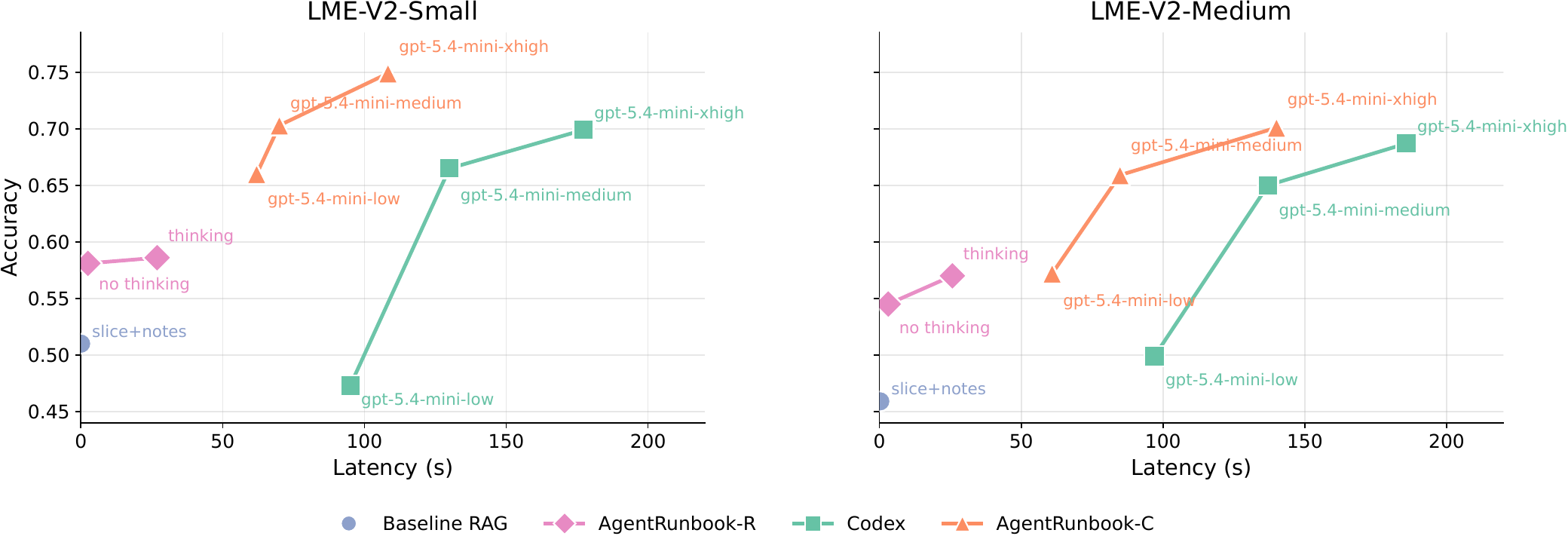}
    \caption{AgentRunbook improves the query accuracy-latency frontier of memory modules.}
    \label{fig:acc-latency-tradeoff}
\end{figure*}

An ideal memory method should support both accurate and efficient querying. Across multiple factors, we find the reasoning effort of the memory controller has a large and direct effects on the overall query latency. We thus use it to analyze the methods across different operating points. As shown in \Cref{fig:acc-latency-tradeoff},  AgentRunbook-R provides a moderate-accuracy, low-latency baseline: it substantially improves over the slice-plus-note baseline while keeping latency around 26 seconds, or much lower without thinking. This makes it a strong choice when query efficiency is prioritized.  AgentRunbook-C moves the accuracy and latency frontier upward. Across reasoning effort settings, the scaffolded coding agent memory consistently offers a better trade-off than directly using the off-the-shelf coding agent. This suggests that coding agents are more effective as memory controllers when paired with explicit workflow guidance, manifests, and trajectory inspection tools.

\section{Conclusion}

We introduce LongMemEval-V2, a long-term memory benchmark that formulates a new standard for agent memory evaluation: memory systems should help agents become experienced operators of specialized environments. LME-V2 holistically covers five memory abilities and advances the context depth of memory benchmarks with beyond 100M-token context from large multimodal web-agent histories. We further propose AgentRunbook-R, which improves standard RAG-based methods with dedicated memory pools, and AgentRunbook-C, which leverages the file manipulation abilities of coding agents and further improves the accuracy and latency through lightweight workflow guidance, manifests, and inspection tools. We hope LME-V2 provides a concrete testbed for memory modules that make long-running agents more intelligent and reliable in real-world environments.

\bibliographystyle{plainnat}
\bibliography{example_paper}

\newpage
\appendix
\onecolumn
\section{LongMemEval-V2: Further Benchmark Construction Details}
\label{appendix-further-benchmark-details}

\subsection{Trajectory Collection}

\paragraph{Domain and task selection.}
We collect trajectories from WebArena, WorkArena, and WorkArena++. For WebArena, we use the OneStopShop, CMS, and Reddit tasks because our pilot inspection found that these websites contain more environment-specific customizations than Wikipedia, Map, and GitLab. WorkArena and WorkArena++ are both based on ServiceNow, so we include all tasks from both benchmarks.

\paragraph{AgentLab harness and action space.}
We use AgentLab to collect trajectories under a unified web agent interface. AgentLab provides benchmark wrappers, observation preprocessing, high-level browser actions, and trajectory logging. The agent observes accessibility tree representations and screenshots, and interacts with the browser through a high-level BrowserGym action API. Elements are referenced by \texttt{bid} identifiers from the accessibility tree, and the agent emits exactly one Python-like action call at each step.

For WorkArena and WorkArena++, we use the following action space:
\[
\begin{gathered}
\texttt{noop(wait\_ms=1000)},\quad
\texttt{scroll(delta\_x, delta\_y)},\quad\\
\texttt{fill(bid, value, enable\_autocomplete\_menu=False)},\\
\texttt{select\_option(bid, options)},\quad
\texttt{click(bid, button='left', modifiers=[])},\quad \\
\texttt{dblclick(bid, button='left', modifiers=[])},\\
\texttt{hover(bid)},\quad
\texttt{press(bid, key\_comb)},\quad
\texttt{focus(bid)},\quad
\texttt{clear(bid)},\\
\texttt{drag\_and\_drop(from\_bid, to\_bid)},\quad
\texttt{tab\_focus(index)},\quad
\texttt{new\_tab()},\quad
\texttt{tab\_close()},\\
\texttt{go\_back()},\quad
\texttt{go\_forward()},\quad
\texttt{goto(url)},\quad\\
\texttt{send\_msg\_to\_user(text)},\quad
\texttt{report\_infeasible(reason)}.
\end{gathered}
\]
For WebArena, we use:
\[
\begin{gathered}
\texttt{noop(wait\_ms=1000)},\quad
\texttt{scroll(delta\_x, delta\_y)},\quad
\texttt{keyboard\_press(key)},\\
\texttt{click(bid, button='left', modifiers=[])},\quad\\
\texttt{fill(bid, value, enable\_autocomplete\_menu=False)},\quad
\texttt{hover(bid)},\\
\texttt{tab\_focus(index)},\quad
\texttt{new\_tab()},\quad
\texttt{go\_back()},\quad
\texttt{go\_forward()},\quad
\texttt{goto(url)},\\
\texttt{tab\_close()},\quad
\texttt{select\_option(bid, options)},\quad\\
\texttt{send\_msg\_to\_user(text)},\quad
\texttt{report\_infeasible(reason)}.
\end{gathered}
\]
The main difference is that WebArena includes \texttt{keyboard\_press(key)}, while WorkArena and WorkArena++ include additional element-specific actions such as \texttt{dblclick}, \texttt{press}, \texttt{focus}, \texttt{clear}, and \texttt{drag\_and\_drop}.

\paragraph{Trajectory collection agents.}
Most trajectories are collected with AgentLab's generic ReAct-style agent using GPT-5-mini and GPT-5.2 as the underlying LLMs. In addition, we use a manual action agent controlled by Codex with GPT-5.2 at xhigh reasoning effort. The manual action agent does not change the environment interface or action space. It displays the current observation, accepts a proposed high-level action string from the operator, validates it against the same BrowserGym action API, and logs the resulting transition in the standard AgentLab trajectory format. We use this Codex-controlled manual-action setup mainly for Level-3 WorkArena++ tasks, where additional targeted collection is needed to obtain successful trajectories across task categories.

\paragraph{Rejection sampling and trajectory filtering.}
Our primary goal during rejection sampling is to obtain successful trajectories for each task, while also retaining useful failure trajectories. We initially aim to collect both one successful and one failed trajectory per task instance when possible. In practice, many tasks end up with only one successful or one failed trajectory because additional sampling becomes costly. The final trajectory pool contains 599 WebArena trajectories and 941 WorkArena/WorkArena++ trajectories. On average the trajectories contain 28.1 states. Overall, 73.1\% of trajectories come from GPT-5-mini, 22.9\% from GPT-5.2, and 4.0\% from Codex with GPT-5.2 (xhigh reasoning).

We also considered trajectories released through the official WebArena leaderboard, but did not use them as the main data source. Some public trajectories lack screenshots, which are essential for our multimodal haystack design. In addition, many leaderboard trajectories are produced by agents specialized for WebArena and are often very short, making them less natural for studying environment experience accumulated from general agents. By contrast, trajectories generated by GPT-5-mini and GPT-5.2 use strong general models and provide successful examples across all selected task categories. We use Codex sparingly because it is more expensive, and because the coding agent setup may inadvertently bring in additional external knowledge from e.g., web searches.

\paragraph{Final trajectory artifact and goal sanitization.}
For each retained trajectory, the final artifact consists of the task goal and an ordered sequence of state-action pairs. Each state contains the screenshot and the accessibility tree, and each action is the high-level BrowserGym action emitted at that step. For WorkArena and WorkArena++ Level-2 and Level-3 tasks, the original goal descriptions sometimes contain detailed navigational hints, which can make procedure questions answerable by reading the route directly from the goal rather than learning it from the trajectory. We therefore sanitize the goal descriptions while preserving the task intent and task-specific payload. We also sanitize repeated appearances of the original goal in trajectory states, such as the initial task description and later copied task text. \Cref{tab:goal_sanitization_examples} shows representative examples.

\begin{table*}[t]
\centering
\small
\setlength{\tabcolsep}{4pt}
\renewcommand{\arraystretch}{1.08}
\caption{Examples of goal sanitization for WorkArena and WorkArena++ Level-2/3 trajectories. The rewritten goals preserve task intent and task-specific values while removing explicit navigation routes and step-by-step module hints.}
\label{tab:goal_sanitization_examples}
\begin{tabular}{p{0.22\linewidth}p{0.36\linewidth}p{0.36\linewidth}}
\toprule
\textbf{Task family} & \textbf{Original goal excerpt} & \textbf{Sanitized goal excerpt} \\
\midrule
Duplicate problem cleanup &
Clean-up your duplicate problems. Concretely, navigate to the ``Assigned to me'' module of the ``Problem'' application. Create a filter where ``Problem statement'' contains \texttt{\#SERIES-5ea261ef-8}. Mark problems with duplicated problem statements as such. &
Clean-up your duplicate problems. Review your own assigned problems where ``Problem statement'' contains \texttt{\#SERIES-5ea261ef-8}. Mark problems with duplicated problem statements as such. \\
\midrule
Dashboard-driven catalog restocking &
Retrieve information from the chart with title \texttt{\#CAT012007808}. Navigate to \texttt{Reports > View/Run}, search for the report, then navigate to \texttt{Self-Service > Service Catalog} and place an order for the least available item. &
Retrieve information from the chart with title \texttt{\#CAT012007808}. Find the greatest stock value and the least available item in the chart. For the least available item, place an order for extra items such that its quantity matches the value you found. \\
\midrule
Requested-item reorder &
Order same item as Kathryn-Lisa Ibarra-Stewart. Navigate to the ``Requested Items'' module of ``Self-Service'', filter by ``Requested for'', then navigate to the ``Service Catalog'' module and order the item with the specified quantity and configuration. &
Order same item as Kathryn-Lisa Ibarra-Stewart. Find the item previously requested for Kathryn-Lisa Ibarra-Stewart. Order the item with the specified quantity and configuration. \\
\midrule
Hardware asset lookup &
Find the warranty expiration date for Julia-Dylan Ray-Mclean's laptop. Navigate to \texttt{Portfolios > Hardware Assets}, filter where ``Assigned to'' is Julia-Dylan Ray-Mclean, and extract the ``Warranty expiration'' field. &
Find the warranty expiration date for Julia-Dylan Ray-Mclean's laptop and report it. \\
\bottomrule
\end{tabular}
\end{table*}

\subsection{Question Annotation}

The annotation team consists of one graduate student and three undergraduate students. Before writing questions, annotators familiarize themselves with the target environments by interacting with the sandbox websites and inspecting collected trajectories. The project lead explains the memory ability definitions as we defined in the main text and annotators discuss ambiguous cases with the project lead throughout the process.

Annotators first inspect trajectories to identify environment-specific facts, state changes, workflows, and gotchas that an experienced colleague should know. They then write questions whose answers require this experience rather than general knowledge of the public base platforms. Each question is validated by at least one additional annotator, who checks that the question is answerable from the intended trajectory evidence, that the answer is correct, and that the question type matches the intended memory ability. When uncertainty remains, the annotators resolve it through discussion with the project lead.

We use the LMArena webpage\footnote{\url{https://arena.ai/}} to test whether strong LLMs can answer candidate questions from parametric knowledge alone. Empirically, we find that models often answer many candidate questions correctly because they know the public versions of Magento, Postmill, and ServiceNow. We therefore filter many questions and rewrite or perturb others until the models fail. In some cases, we first write a free-form question, observe the models' incorrect answers, and then use these wrong answers as distractors to construct a more challenging multiple-choice question. In the final context gathering evaluation, the reader is also prompted to answer based on the returned memory context and to avoid guessing, which further reduces the chance that accuracy comes from parametric knowledge rather than retrieved experience.

For all questions, annotators verify that the relevant answer evidence is visible in screenshots. Although accessibility trees sometimes expose information more clearly than screenshots, especially for hidden labels or structured fields, we avoid questions whose answer can only be inferred from the accessibility tree and not from the visual trajectory evidence.

\subsection{Answer Trajectory Labeling} 

During question annotation, annotators record a seed set of trajectories that contain the answer for each question. However, the final haystacks are shared across questions, so we perform an additional answer-trajectory labeling pass to identify all trajectories that contain sufficient evidence for each question. The output of this step is a coverage map from each question ID to the set of answer-bearing trajectory IDs. For questions that require multiple independent pieces of evidence, such as cross-form comparisons or gotchas involving both symptom and resolution, the coverage map also records required evidence hops. The haystack builder later enforces that at least one trajectory is selected for each required hop.

We use Codex to assist this labeling process. Each worker is assigned a small batch of questions together with the relevant trajectory pool, metadata, and workflow rules. We found batching important for accuracy: it lets the agent focus on a coherent question family, such as static form comparison, dynamic transition, workflow, or gotcha questions, while keeping the search space small enough for careful inspection. The workflow emphasizes both recall and precision. The agent first narrows candidate trajectories using metadata, task family, URL patterns, and trajectory summaries, and then inspects candidate trajectories directly. Inclusion requires screenshot-visible evidence; accessibility trees and string matching can be used for triage, but they are not sufficient for inclusion. Failure trajectories are treated equally with successful trajectories, since many questions are answerable only from failed or partially completed runs. When the evidence is ambiguous, the agent records the trajectory as uncertain with a rationale rather than forcing an inclusion decision.

The workflow also enforces question-type-specific rules. Static questions require direct visual evidence of the relevant page, field, value, or form. Dynamic questions require evidence of the relevant before/after state change rather than only a related task family. Procedure questions require the trajectory to show the key procedural steps, not merely a similar entity or goal. Gotcha questions are marked as high-risk by default because the required evidence is often subtle and context-dependent. For multi-hop questions, each hop must have its own candidate trajectory set, and a single trajectory may satisfy multiple hops only when the cited screenshots directly support each hop.

After the Codex-assisted pass, humans manually resolve the ambiguous cased marked by the agent and manually validate the question-trajectory correspondence for the trajectories included in the final core haystack set. Validators check that each selected trajectory contains the intended evidence, that the evidence is visible in screenshots, and that multi-hop constraints are satisfied. This final human validation is used to ensure that the minimal answer core used in haystack construction preserves answer coverage while avoiding spurious answer-bearing trajectories.

\subsection{Haystack Creation}

We construct haystacks in three stages: answer-core selection, small-haystack expansion, and medium-haystack expansion. The input to the builder is the final coverage map from the previous step, which lists all answer-bearing trajectories for each question. Some questions require multiple evidence hops, so the coverage map can specify multiple required hops, each with its own candidate trajectory set. Abstention questions with no direct answer-bearing trajectory are handled separately through a manually verified anchor mapping to the corresponding source question.

\paragraph{Minimal answer core.}
We first build a minimal answer core for each domain. The goal is to select a small shared set of trajectories such that every nonzero-support question has at least one selected answer-bearing trajectory for each required hop. We formulate this as an assignment problem: for each question-hop requirement, the program chooses one trajectory from its candidate answer-bearing set. The objective favors trajectories that cover more questions and more requirements, while also encouraging sparse coverage, avoiding unnecessary overcoverage, controlling the success/failure ratio, and minimizing the number of unique selected trajectories. We solve the assignment with deterministic greedy initialization and local search over multiple restarts. The resulting core contains 44 trajectories for WebArena and 49 trajectories for ServiceNow. We then manually verify the minimal cores to ensure that the selected trajectories contain the intended evidence for the questions and that multi-hop requirements are represented.

\paragraph{Small haystack.}
Starting from the verified minimal core, we expand each domain to a shared 100-trajectory haystack. The WebArena questions share one 100-trajectory haystack, and the ServiceNow questions share another. Fillers are selected from trajectories outside the minimal core. The filler ranking encourages diversity over task families, prefers trajectories with low global answer-support count, applies a weak preference for harder trajectories, and controls the success/failure balance toward a 1:1 ratio. After construction, the trajectory order is deterministically shuffled. 

\paragraph{Medium haystack.}
The medium tier is built per question rather than as a single shared haystack. Importantly, it reuses the same per-question answer seed selected by the small-haystack builder, so the answer-bearing core does not change between tiers. For each question, the builder starts from the selected answer trajectories for its required hops and then adds fillers until the target size is reached, using 500 trajectories for the reported medium tier. Fillers are sampled after excluding the full answer-bearing set for that question, ensuring that the expansion does not introduce extra answer trajectories beyond the chosen seed. The filler ranking promotes task-family diversity, low overlap between filler goals and answer-core goals, low global answer-support count, and outcome balance. For zero-support abstention questions, the builder reuses the haystack of the manually verified anchor question exactly.

\subsection{Evaluation} 
We evaluate each memory system with the context gathering protocol. For each question, the evaluation harness first inserts all trajectories in the question's haystack into the memory module, queries the final memory with the question, validates that the returned context consists only of text and image items, and truncates the context to 200K Qwen3.5-9B tokens. The fixed reader then receives the domain-specific system prompt, the returned memory context, the question text, and the question image if present. The reader output is parsed by extracting the last \texttt{\textbackslash boxed\{\}} expression; if no boxed answer is found, the full response is used as the prediction. A prediction of \texttt{UNKNOWN} is always counted as incorrect for accuracy.

Structured answers are scored with deterministic evaluators specified by each question, including normalized phrase matching, ordered phrase matching, single-choice matching, and multi-select choice matching. Gotchas and abstention questions are evaluated with an LLM judge because they require semantic comparison to the reference insight or flawed-premise explanation. The judge outputs a binary JSON label, which is used as the final score. We aggregate accuracy over the full set and also report category-level breakdowns.

\paragraph{Experimental Details} For the fixed reader model, we use the Qwen3.5-9B model hosted with vllm \citep{vllm} on a local machine with Nvidia A100 GPUs. We use temperature 0.6, top\_p 0.95, and top\_k 20 for a sampling evaluation. We perform manual prompt tuning to make sure that the prompt template is reliable such that (1) with no memory context the model's performance is near zero and (2) with the oracle image+text state slices, the model's performance is as high as possible. For evaluating the correctness for abstention and gotchas questions we use the GPT-5.2 model with medium reasoning effort as the LLM judge. 

\begin{table*}[t]
\centering
\small
\setlength{\tabcolsep}{4.0pt}
\renewcommand{\arraystretch}{1.08}
\caption{Reader prompt used in the main context gathering evaluation. The same template is used for all memory systems.}
\label{tab:reader_prompt}
\begin{tabular}{p{0.18\linewidth}p{0.77\linewidth}}
\toprule
\textbf{Component} & \textbf{Template} \\
\midrule
System prompt, WebArena &
You are an experienced colleague in a web browsing environment that has a customized Magento-based shopping website, a customized Magento-based shopping admin CMS website, as well as a customized forum website based on Reddit/Postmill. Answer based on your memory of the environment. If you do not know the answer, output exactly \texttt{\textbackslash boxed\{UNKNOWN\}}. Do not guess. Never attempt to guess an answer if you are not sure. If you believe the question's construction/premise is wrong, provide an explanation in \texttt{\textbackslash boxed\{\}} explaining why the question is flawed. \\
\midrule
System prompt, WorkArena &
You are an experienced colleague working in a customized ServiceNow environment. Answer based on your memory of the environment. If you do not know the answer, output exactly \texttt{\textbackslash boxed\{UNKNOWN\}}. Do not guess. Never attempt to guess an answer if you are not sure. If you believe the question's construction/premise is wrong, provide an explanation in \texttt{\textbackslash boxed\{\}} explaining why the question is flawed. \\
\midrule
User prompt &
\texttt{\#\#\# Memory context:} \newline
\texttt{\{memory context returned by the memory module, after truncation\}} \newline
\texttt{\#\#\# Question to answer:} \newline
\texttt{\{question text\}} \newline
\textit{If the question contains an image, the image is appended after the question text.} \\
\bottomrule
\end{tabular}
\end{table*}

\begin{table*}[t]
\centering
\small
\setlength{\tabcolsep}{4.0pt}
\renewcommand{\arraystretch}{1.08}
\caption{LLM-judge prompts for semantic evaluation of abstention and gotchas questions. Both judges return JSON with a binary label.}
\label{tab:llm_judge_prompts}
\begin{tabular}{p{0.16\linewidth}p{0.79\linewidth}}
\toprule
\textbf{Judge} & \textbf{Prompt} \\
\midrule
Abstention judge &
\textbf{System:} You are a strict grader for flawed-premise (abstention) questions. Judge whether a model answer correctly identifies that the question premise is wrong, consistent with the reference answer. If the model follows the flawed premise and gives a concrete answer under that premise, it must be graded 0. If the model's final answer is just UNKNOWN / cannot determine without identifying the flaw, grade 0. If the model is contradictory (both rejects premise and also gives a concrete premise-following answer), grade 0. Paraphrases are allowed when they preserve the same core flaw described by the reference answer.

\textbf{User:} Evaluate whether the model correctly handled a flawed-premise question.

\textbf{Question:} \texttt{\{question\_text\}}

\textbf{Reference answer:} \texttt{\{reference\_answer\}}

\textbf{Model full response:} \texttt{\{model\_full\_response\}}

\textbf{Model extracted final answer:} \texttt{\{model\_final\_answer\}}

\textbf{Scoring rubric:} Label 1 only if the model clearly identifies the flawed premise and reaches a conclusion equivalent to the reference answer. Label 1 also if the model explicitly says it lacks access to the user's specific live environment/instance/configuration and therefore cannot verify, provided it does not give a concrete premise-following answer. Label 0 if the model follows the flawed premise and gives a concrete answer under that premise. Label 0 for generic UNKNOWN/insufficient-info replies that do not identify a flaw and do not make the explicit environment-access limitation clear. Label 0 if contradictory.

\textbf{Output JSON only:} \texttt{\{"label": 0 or 1, "reason": "short rationale"\}} \\
\midrule
Gotchas judge &
\textbf{System:} You are a strict grader for gotchas-style insight questions. The reference answer describes the key insight(s). Grade 1 if the model response includes at least one correct insight point from the reference answer (paraphrase allowed), and does not contradict any reference point. If the model's direction is wrong, or it contains contradictions against any reference point, grade 0. If the model gives multiple points, partial coverage is enough for 1 as long as no contradictions appear.

\textbf{User:} Evaluate whether the model answer captures the gotcha insight.

\textbf{Question:} \texttt{\{question\_text\}}

\textbf{Reference answer:} \texttt{\{reference\_answer\}}

\textbf{Model full response:} \texttt{\{model\_full\_response\}}

\textbf{Model extracted final answer:} \texttt{\{model\_final\_answer\}}

\textbf{Scoring rubric:} Label 1 if the model includes at least one correct insight point from the reference answer (paraphrase acceptable), and does not contradict any reference point. Label 1 even if only part of a multi-point reference answer is covered, as long as there is no contradiction. Label 0 if direction is wrong (suggests opposite action/cause), even if some wording overlaps. Label 0 if any point in the model response contradicts any reference point. Label 0 if the response is irrelevant or generic without insight.

\textbf{Output JSON only:} \texttt{\{"label": 0 or 1, "reason": "short rationale"\}} \\
\bottomrule
\end{tabular}
\end{table*}

\section{LongMemEval-V2: Pilot Studies}
\label{appendix-pilot-study}

We conduct two pilot studies to understand whether the benchmark questions can be answered without trajectory evidence and whether oracle access to answer-bearing trajectories is sufficient for reliable question answering. These studies use a direct question answering setup rather than the context gathering formulation used in the main experiments. We report results on non-abstention questions only, since abstention questions are intentionally constructed with misleading premises and do not primarily test ordinary evidence use.

\subsection{Can Frontier Models Answer Without Trajectory History?}
\label{appendix-pilot-study-no-context}

We first evaluate whether recent frontier models can answer LME-V2 questions from parametric knowledge alone. Each model receives only the question and, when applicable, the question image. The model is instructed to answer directly and to output \texttt{\textbackslash boxed\{UNKNOWN\}} rather than guessing. We use OpenRouter \citep{openrouter2026} to perform the evaluation and medium reasoning effort is used across all models. As shown in \Cref{tab:pilot_no_context}, all models perform poorly without trajectory evidence, with the best overall accuracy reaching only 14.1\%. This confirms that the questions generally depend on environment-specific experience rather than public or parametric knowledge. In other words, although these models have substantial knowledge of the public versions of the environments (Magento, Postmill, ServiceNow, etc.), they still lack the environment-specific knowledge to enable them as experienced colleagues in the WebArena and WorkArena websites. 

\begin{table}[t]
\centering
\small
\setlength{\tabcolsep}{4.2pt}
\renewcommand{\arraystretch}{1.08}
\caption{No-context direct QA results on non-abstention questions. Frontier LLMs perform poorly across problem types.}
\label{tab:pilot_no_context}
\begin{tabular}{lccccc}
\toprule
\textbf{Model} & \textbf{Overall} & \textbf{Static} & \textbf{Dynamic} & \textbf{Workflow} & \textbf{Gotchas} \\
\midrule
GPT-5.2 & 0.047 & 0.000 & 0.000 & 0.032 & 0.210 \\
Gemini-3.1-Pro-Preview & 0.110 & 0.104 & 0.096 & 0.147 & 0.241 \\
Claude Opus 4.6 & 0.118 & 0.096 & 0.121 & 0.134 & 0.379 \\
GLM-5V-Turbo & 0.101 & 0.126 & 0.107 & 0.091 & 0.205 \\
Grok-4.20 & 0.024 & 0.000 & 0.029 & 0.151 & 0.102 \\
Kimi-K2.5 & 0.141 & 0.183 & 0.197 & 0.115 & 0.171 \\
Qwen3.6-Plus & 0.110 & 0.118 & 0.078 & 0.091 & 0.310 \\
\bottomrule
\end{tabular}
\end{table}

\subsection{Can Models Reliably Answer with Oracle Trajectory Access?}
\label{appendix-pilot-study-oracle-context}

We next evaluate whether models can answer when given oracle access to the trajectories that contain the answer. This setting removes the retrieval problem and isolates the difficulty of reading, grounding, and reasoning over trajectory evidence. We compare three variants. First, \textit{oracle trajectories} provides the full answer-bearing trajectories. Second, \textit{oracle slices + notes} provides procedure and hint notes together with radius-1 evidence windows around the annotated answer states. The notes are generated using a reflection prompt conditioned on a compressed view of the trajectory's content. Third, we evaluate a coding agent direct QA setup, where Codex explores a local sandbox containing the oracle trajectories and writes the answer to a JSON file. The Codex runs use Codex binary v0.117.0 with \texttt{gpt-5.4-mini (xhigh)}.

\Cref{tab:pilot_slice_note_prompt} shows the prompt template used for direct QA with rendered oracle trajectory context. The data-dependent trajectory content is abbreviated. For the coding agent experiments, we package each question as an isolated local sandbox. The model is not asked to return a memory context. Instead, it directly answers the question by inspecting local files and writing the result to \texttt{answer.json}. \Cref{tab:pilot_codex_prompt} summarizes the layout and instruction.

The \Cref{fig:pilot_studies} right table shows two main findings. First, full oracle trajectories are not sufficient for reliable direct QA: Qwen3.5-9B reaches 59.6\% and GPT-5.4-mini (medium) reaches 65.3\%. Second, evidence slicing and notes substantially improve direct QA, reaching 82.5\% and 86.3\%, respectively. Finally, Codex direct QA further improves to 89.7\%, suggesting that file-system exploration and iterative evidence inspection are effective ways to process agent trajectories. With inherent file system manipulation and tool use capabilities, general coding agents promise as effective memory controllers, motivating our AgentRunbook-C memory module design.

\begin{table*}[t]
\centering
\small
\caption{Prompt template for oracle slices and notes direct QA. The same template is used for full oracle trajectories.}
\label{tab:pilot_slice_note_prompt}
\begin{tabular}{p{0.17\linewidth}p{0.78\linewidth}}
\toprule
\textbf{Component} & \textbf{Template} \\
\midrule
System prompt, web &
You are an experienced colleague in a web browsing environment that has a customized Magento-based shopping website, a customized Magento-based shopping admin CMS website, as well as a customized forum website based on Reddit/Postmill. Answer based on your memory of the environment. If you do not know the answer, output exactly \texttt{\textbackslash boxed\{UNKNOWN\}}. Do not guess. Never attempt to guess an answer if you are not sure. If you believe the question's construction/premise is wrong, provide an explanation in \texttt{\textbackslash boxed\{\}} explaining why the question is flawed. \\
\midrule
System prompt, ServiceNow &
You are an experienced colleague working in a customized ServiceNow environment. Answer based on your memory of the environment. If you do not know the answer, output exactly \texttt{\textbackslash boxed\{UNKNOWN\}}. Do not guess. Never attempt to guess an answer if you are not sure. If you believe the question's construction/premise is wrong, provide an explanation in \texttt{\textbackslash boxed\{\}} explaining why the question is flawed. \\
\midrule
User prompt &
\texttt{\# Memory context:} \newline
\texttt{\#\# Procedure and Hint Notes Learned from Previous Tasks in the Environment} \newline
\textit{For each oracle trajectory: procedure note title, description, and bullet content; hint note title, description, and bullet content.} \newline
\texttt{\#\# Oracle Trajectories and Relevant State Slices from Previous Tasks in the Environment} \newline
\textit{For each selected trajectory: goal, outcome, start URL, action list, and evidence windows centered at annotated answer states. Each evidence window includes states from radius 1 around the annotated state, with URL, action, accessibility-tree text, and screenshots according to the rendering configuration.} \newline
\texttt{\# Question to answer:} \newline
\texttt{\{question\}} \newline
\textit{If the question contains an image, the image is appended after the question text.} \\
\bottomrule
\end{tabular}
\end{table*}

\begin{table*}[t]
\centering
\small
\caption{Sandbox layout and instruction for the Codex oracle direct-QA pilot study.}
\label{tab:pilot_codex_prompt}
\begin{tabular}{p{0.19\linewidth}p{0.76\linewidth}}
\toprule
\textbf{Component} & \textbf{Content} \\
\midrule
Sandbox layout &
\texttt{question.json}: question text and optional copied question image. \newline
\texttt{INSTRUCTION.md}: task instruction. \newline
\texttt{answer.json}: initialized as \texttt{\{"answer": ""\}}. \newline
\texttt{trajectories/\{trajectory\_id\}/trajectory.json}: oracle trajectory with \texttt{id}, optional \texttt{original\_goal}, optional \texttt{outcome}, and state content. \newline
\texttt{trajectories/\{trajectory\_id\}/screenshots/}: copied trajectory screenshots. \\
\midrule
Package instruction &
You are an experienced colleague working in a customized web environment. Read \texttt{question.json} and inspect every trajectory under \texttt{trajectories/}. This package comes from a public-environment-based setup that has been customized. Do not rely on prior knowledge of the public environment. Work only from the provided trajectories and copied question assets. Use only the question and the provided trajectories. Do not browse anywhere else, do not inspect other question folders, and do not use outside resources. If the question specifies an answer format, follow it exactly. For multiple-choice questions, write only the boxed letter corresponding to your answer, e.g., \texttt{\textbackslash boxed\{A\}}, into the answer field. Write your final answer to \texttt{answer.json} using this exact schema: \texttt{\{"answer": "<your final answer>"\}}. \\
\midrule
Codex invocation prompt &
You are an experienced colleague working in a customized web environment. Read the local files in this directory, especially \texttt{INSTRUCTION.md} and \texttt{question.json}. If \texttt{question.json} refers to a screenshot, view it carefully. Use only local files in this directory. Solve the task and write your final answer to \texttt{answer.json} as valid JSON with a non-empty string field named \texttt{answer}. If \texttt{answer.json} already exists, update only the \texttt{answer} value. Follow the formatting instructions in \texttt{question.json}. For multiple-choice questions, write only the boxed letter corresponding to your answer, e.g., \texttt{\textbackslash boxed\{A\}}, into the answer field. \\
\bottomrule
\end{tabular}
\end{table*}

\section{Implementation Details}
\label{app:agentrunbook_details}
\label{section-impl-details}

\subsection{Baselines}
\label{section-impl-details-baselines}

Since most existing memory systems mainly consider the conversation as the main input modality and facts as the corresponding memory value representations, LME-V2 poses a significant challenge and a direct naive adaptation would lead to suboptimal performance. As a result, we mainly consider two RAG and coding agent methods that we found strongest during our pilot study during our development of AgentRunbook, as well as studying ablation versions of our proposed system.

\paragraph{Slice+Note RAG} Based on our pilot studies in \Cref{appendix-pilot-study}, a RAG-based memory system have strong performance if its state recall is accuracy and combines high-level insights with the low-level details. We thus inherit from this design and build two separate knowledge pools: the slice pool and the high-level note pool. We use the same hyperparameters as AgentRunbook-R to construct both pools. For the hyperparameters, we conducted a grid search for slice radius size (value 1, 2, 3), slice modality (image only, axtree text only, image+axtree), and top-k (3, 6, 9). We find that radius 1 with image+axtree slices and k=6 works the best. At query time, the question itself is directly used as the query for both pools. We retain top-6 states and top-3 notes. Qwen3-8B-Embedding is used as the embedding for all the experiments. 

\paragraph{Codex Agent} As our pilot study found promising performance of off-the-shelf coding agents, we include Codex as the baseline following the same sandbox setting we used for the pilot study. We mainly used Codex due to its performance, popularity, and its availability as an open-source project. Codex v0.117.0 binary is used for all the experiments. We download the software from github release\footnote{We use the binary \url{https://github.com/openai/codex/releases/download/rust-v0.117.0/codex-x86_64-unknown-linux-musl.tar.gz}.} and run from a local linux server. The server has the common software required by Codex, especially \texttt{ripgrep} and \texttt{find}.  In the early stages of the project, we manually tuned the instruction for this baseline to make sure that it is clear and informative so that the evaluation results do not underrepresent the agent's ability. We use GPT-5.4-mini due to its competitive latency and cost. 

\subsection{AgentRunbook-R}
\label{app:agentrunbook_r_details}

AgentRunbook-R maintains three retrieval pools. For each inserted trajectory, we first materialize a simplified trajectory containing state indices, URLs, actions, accessibility-tree text, screenshots, and task metadata. The raw-state pool contains one entry per state. Each entry stores a radius-1 local state window, the full trajectory action sequence, the local action sequence, the trajectory goal, and the center-state screenshot. The event pool contains one entry per adjacent state transition. Each event is generated by the memory controller LLM from the pre-state, post-state, and annotated action trace, and contains a concise transition title and description. The note pool contains two trajectory-level entries: one procedure note and one hint note. The procedure note summarizes reusable task steps, while the hint note records durable environment-specific observations and pitfalls.

All pool entries are embedded when the trajectory is inserted. At query time, the memory controller LLM reasons about the current memory state and the query and emits a structured JSON query bundle:
\[
\{
\texttt{"raw\_state\_queries"}: [\ldots],
\texttt{"event\_query"}: \ldots,
\texttt{"note\_query"}: \ldots
\}.
\]
The raw-state field may contain up to five focused queries, each targeting a distinct UI surface or entity. The event query targets interaction effects, navigation changes, popups, blockers, and other state transitions. The note query targets reusable procedures, module paths, disambiguation rules, and gotchas. AgentRunbook-R retrieves the top entries from each pool with dense similarity search, merges duplicate raw-state results, and renders the retrieved notes, events, and raw state slices in this order. The rendered context includes text and screenshots when available.

\paragraph{Prompt templates.}
\Cref{tab:agentrunbook_r_prompts} summarizes the fixed prompts used by AgentRunbook-R. 

\paragraph{Experimental Details} For the main experiments in the paper, we limit the number of raw state queries to 5. We maintain top-6 events and top-3 notes. As the number of queries for the raw states is flexible, we retain top-\texttt{m} results for each query, where \texttt{m=min(2,6//n\_queries)}. We use Qwen3-Embedding-8B model for all the retrieval purposes. The input is truncated to 4,096 tokens. We use Qwen3.5-9B as the memory controller. We temperature 0.6, top\_p 0.95 across settings. For the index-time knowledge pool generation, we always enable thinking. For query-time query generation, we experimented with both thinking enabled and disabled and report both in \Cref{fig:acc-latency-tradeoff}. For the main results in \Cref{tab:main_results} the results are measured with thinking enabled. All the embedding and generation models are served with vllm on a local machine with Nvidia A100 GPUs.

\begin{table*}[t]
\centering
\small
\setlength{\tabcolsep}{4.0pt}
\renewcommand{\arraystretch}{1.08}
\caption{Prompt templates used by AgentRunbook-R. Trajectory-specific content is abbreviated.}
\label{tab:agentrunbook_r_prompts}
\begin{tabular}{p{0.18\linewidth}p{0.77\linewidth}}
\toprule
\textbf{Prompt} & \textbf{Template} \\
\midrule
Procedure and hint note generation &
\textbf{System:} You convert one UI task trajectory into two reusable memory notes for a future agent. Assume these notes will later be retrieved for unknown future questions. Preserve the workflow and the highest-value reusable facts from the touched pages. Write \texttt{procedure\_note} and \texttt{hint\_note}. Each note must contain \texttt{title}, \texttt{description}, and \texttt{content}. Use only evidence grounded in the provided goal, outcome, thoughts, annotated actions, and screenshots. Do not invent unseen fields, filters, modules, or outcomes. If the run failed, describe only the intended or attempted workflow where the evidence supports it. Keep the procedure note focused on the reliable core workflow and use the hint note for durable facts, pitfalls, option sets, confirmation signals, absent functionality, and distinctions between easily confused controls. Return only valid JSON:
\texttt{\{"procedure\_note":\{"title":"...","description":"...","content":"- ..."\},"hint\_note":\{"title":"...","description":"...","content":"- ..."\}\}}.

\textbf{User:} Extract two reusable notes from this UI task run. Goal: \texttt{\{goal\}}. Outcome: \texttt{\{outcome\}}. Start URL: \texttt{\{start\_url\}}. Each state block is followed by the screenshot for that state. The action line is the action taken from that state, annotated with recoverable object or module details. \textit{Then include the ordered state blocks, thoughts, annotated actions, and screenshots.} \\
\midrule
State-transition event generation &
\textbf{System:} You convert one UI transition from a longer task trajectory into retrieval-ready event text. You will be given the task goal and outcome, the full annotated action trace, and one target transition defined as pre-state, annotated action, and post-state. Return exactly one JSON object:
\texttt{\{"overview":"...","state\_transition":"..."\}}. The overview briefly recaps the task goal and workflow stage. The state-transition field explicitly compares the post-state to the pre-state and describes what changed after the action, such as a new page, revealed panel, form fields, changed values, confirmation signal, blocker, popup, navigation, or lack of visible change. Ground the output only in the provided evidence and preserve exact UI labels when available.

\textbf{User:} Generate an event for transition \texttt{\{state\_i\} -> \{state\_j\}}. Goal: \texttt{\{goal\}}. Outcome: \texttt{\{outcome\}}. Full action trace: \texttt{\{actions\}}. Pre-state: \texttt{\{url, thoughts, action, AXTree, screenshot\}}. Post-state: \texttt{\{url, thoughts, action, AXTree, screenshot\}}. \\
\midrule
Query generation &
\textbf{System:} You generate structured retrieval queries for an active memory system with three pools: raw state slices, state-transition events, and procedure/hint notes. Return exactly one JSON object:
\texttt{\{"raw\_state\_queries":["..."],"event\_query":"...","note\_query":"..."\}}. Maximize retrieval of memory entries that would help answer the question later. Do not answer the question yourself. Use raw-state queries for exact UI surface evidence, such as pages, forms, records, tabs, fields, buttons, dropdowns, options, labels, values, counts, and missing controls. Use the event query only when navigation, before/after change, revealed content, confirmation, blocker, popup, or workflow stage matters. Use the note query for reusable procedures, module paths, disambiguation, absent functionality, pitfalls, and durable hints. Remove formatting instructions and final-answer wrappers. Preserve exact entity names and literal UI labels. Deduplicate raw-state queries and cap them at five. Return JSON only.

\textbf{User:} Memory pool summary: \texttt{\{runtime\_summary\}}. Output schema example: \texttt{\{schema\_example\}}. Prompt examples: \texttt{\{few\_shot\_examples\}}. Question ID: \texttt{\{question\_id\}}. Question type: \texttt{\{question\_type\}}. Question text: \texttt{\{question\}}. Question image path: \texttt{\{image\_path or <none>\}}. Original goals attached to this benchmark question: \texttt{\{original\_goals\}}. Return only the JSON object. \\
\bottomrule
\end{tabular}
\end{table*}

\subsection{AgentRunbook-C}
\label{app:agentrunbook_c_details}

AgentRunbook-C uses file storage at insertion time. Each inserted trajectory is stored under the memory workspace as a trajectory directory containing the trajectory JSON and copied screenshots. Query-time retrieval is executed inside an isolated sandbox. The coding agent is instructed to act as a memory retrieval module rather than a final answerer. It inspects local files, identifies compact supporting evidence, and writes a structured memory output.

Before invoking the coding agent, AgentRunbook-C renders two manifest artifacts for the current haystack: a concise trajectory summary and a fuller trajectory summary. These artifacts provide trajectory-level metadata and lightweight previews that help the agent shortlist likely trajectories before detailed inspection. The sandbox also includes a trajectory inspection helper script, which supports targeted operations such as inspecting one trajectory, one state, one state span, or text matches within a trajectory.

The coding agent writes its result to \texttt{memory\_module\_output.json} with the following schema:

\quad\texttt{"memory\_markdown"}: \texttt{string},

\quad\texttt{"trajectory\_spans"}: [
\{\texttt{"trajectory\_id"}: \texttt{string},
\texttt{"start\_state\_index"}: \texttt{int},
\texttt{"end\_state\_index"}: \texttt{int}\}
].

The \texttt{memory\_markdown} field contains brief support analysis and any relevant procedure or hint notes. The trajectory spans are zero-based inclusive state ranges, with a total span budget of 20 states. After the coding agent finishes, AgentRunbook-C validates the JSON output, filters invalid spans, renders the selected states and screenshots into the returned memory context, and passes this context to the fixed reader. 

\paragraph{Sandbox} \Cref{tab:agentrunbook_c_layout} shows the query-time sandbox provided to the coding agent. \Cref{tab:agentrunbook_c_prompts} summarizes the fixed prompt and workflow instruction used by AgentRunbook-C. Long examples and repeated rules are omitted for compactness.

\paragraph{Experimental Details} Codex v0.117.0 binary is used for all the experiments. The query process is implemented by a python harness that prepares the sandbox and invokes \texttt{codex exec} directly. To accelerate evaluation while ensuring a fair latency measurement, we restrict the parallel query invocation to 3 query processes to avoid overloading the disk or network.

\begin{table}[t]
\centering
\small
\setlength{\tabcolsep}{4.2pt}
\renewcommand{\arraystretch}{1.08}
\caption{AgentRunbook-C query-time sandbox layout.}
\label{tab:agentrunbook_c_layout}
\begin{tabularx}{\linewidth}{@{}>{\raggedright\arraybackslash}p{0.37\linewidth}X@{}}
\toprule
\textbf{Path} & \textbf{Content} \\
\midrule
\path{question.json} &
Question text, question id, metadata, and optional copied question image path. \\
\path{INSTRUCTION.md} &
Workflow instruction for using the sandbox as a memory retrieval module. \\
\path{trajectories/} &
Symlink to the inserted trajectory haystack. \\
\path{trajectories/<trajectory_id>/trajectory.json} &
One full trajectory, including goal, start URL, outcome, actions, and ordered states. \\
\path{trajectories/<trajectory_id>/screenshots/} &
Screenshots referenced by trajectory states. \\
\path{trajectories/TRAJECTORY_SUMMARY_CONCISE.md} &
Compact trajectory-level manifest for quick triage. \\
\path{trajectories/TRAJECTORY_SUMMARY_FULL.md} &
Fuller manifest with detailed thought and action traces for shortlist selection. \\
\path{scripts/inspect_trajectory.py} &
Helper script for inspecting one trajectory, state, span, or text match. \\
\path{memory_module_output.json} &
Structured output written by the coding agent. \\
\bottomrule
\end{tabularx}
\end{table}

\begin{table*}[t]
\centering
\small
\setlength{\tabcolsep}{4.0pt}
\renewcommand{\arraystretch}{1.08}
\caption{Prompt and workflow instruction for AgentRunbook-C. Codex invocation
prompt is the prompt for invoking the codex binary software. The other rows in the table are the content in the INSTRUCTION.md.}
\label{tab:agentrunbook_c_prompts}
\begin{tabular}{@{}p{0.18\textwidth}p{0.77\textwidth}@{}}
\toprule
\textbf{Component} & \textbf{Template} \\
\midrule
Codex invocation prompt &
You are acting as the query-time agent for Coding Agent Memory. Read the local files in this directory, especially \texttt{INSTRUCTION.md} and \texttt{question.json}. The local \texttt{trajectories/} directory contains the current haystack for this evaluation item, and you must explore \texttt{trajectories/} before returning your final result. If \texttt{question.json} refers to an image, view it carefully. Write your final result to \texttt{memory\_module\_output.json} as valid JSON. Use the local inspection helper under \texttt{scripts/} when you need to inspect one trajectory, one state, one span, or match text within one trajectory quickly. \\
\midrule
Task overview in \texttt{INSTRUCTION.md} &
You are acting as a quick memory retrieval module to provide contexts from agent trajectories collected from a customized web environment for a downstream reader to answer questions specific to that environment. The question is in \texttt{question.json}. Aggregate information from the local \texttt{trajectories/} directory. Follow the workflow and do not attempt to re-verify or rebuild maps unnecessarily, since the task has latency constraints. Be quick and do not over-explore unless necessary. Work inside the current directory and never explore outside it. \\
\midrule
Output requirement &
Write the final result to \texttt{memory\_module\_output.json} as valid JSON:
\texttt{\{"memory\_markdown":"\#\# Support Analysis\textbackslash n...\textbackslash n\textbackslash n\#\# }\texttt{Relevant Procedure and Hint Notes\textbackslash n...",}\texttt{"trajectory\_spans":[\{"trajectory\_id":"...",}\par \texttt{"start\_state\_index":0,"end\_state\_index":0\}]\}}.
The support analysis should briefly describe where the evidence can be found. If the evidence contradicts the premise of the question, clearly say that the premise is wrong and include the contradicting evidence. The trajectory spans must use zero-based inclusive indices and preserve order by importance. \\
\midrule
Workflow instruction &
First classify the question before opening trajectories in detail. If the question contains an image, inspect it and align it with the matching surface or state. For direct lookup questions, find an exact state showing the requested field, value, button, or page. For comparison questions, find one supporting state per side when needed. For procedure questions, stay within the same workflow family unless the question explicitly asks for a shared pattern across workflows. Start from \texttt{TRAJECTORY\_SUMMARY\_FULL.md} and shortlist only a few likely trajectories using the goal, start URL, action sequence, and final reward. Prefer the helper script for exact verification:
\texttt{python scripts/inspect\_trajectory.py <trajectory\_id>},
\texttt{--state <i>},
\texttt{--span <i:j>}, or
\texttt{--match "<pattern>"}.
Use the helper on shortlisted trajectories rather than performing broad raw-file search. Keep the final evidence package small, usually no more than three states per span, and use at most 20 states in total. \\
\midrule
Final rules &
Move fast and prefer targeted exploration. Put the most important evidence first. Avoid redundant trajectories when multiple trajectories support the same fact. Reject nearby but non-exact matches. Do not copy screenshots or large AXTree blocks into the JSON output. You may write scratch files in the current directory if needed. \\
\bottomrule
\end{tabular}
\end{table*}

\section{Further Analyses}
\label{appendix-further-analysis}

This appendix section provides three analyses of the main methods in \Cref{tab:main_results}. Unless stated otherwise, web and enterprise questions are pooled within each LME-V2 tier.

\subsection{Error Analyses}
We categorize each incorrect final answer into one mutually exclusive error type. Retrieval errors are defined against the final human labeled answer-trajectory coverage maps. For each question, we take the answer-bearing trajectory set and derive each trajectory's task family, which contains the tasks with the same goal but slightly different data and starting points. A major retrieval miss means that the returned raw slices or state spans do not hit any answer-bearing task family. A minor retrieval miss means that the returned evidence hits the answer task family, but not an exact answer-bearing trajectory. If the returned context hits an exact answer-bearing trajectory, answer URL, or gold-answer text, but the reader still answers incorrectly, we label the failure as a reading error. We remark that under this categorization, the memory module often still has the responsibility to the reading errors as it has the freedom to use better state slices and better evidence presentation methods. Gotcha and premise/abstention failures are kept separate for better visualization.

\begin{figure}[t]
\centering
\includegraphics[width=\linewidth]{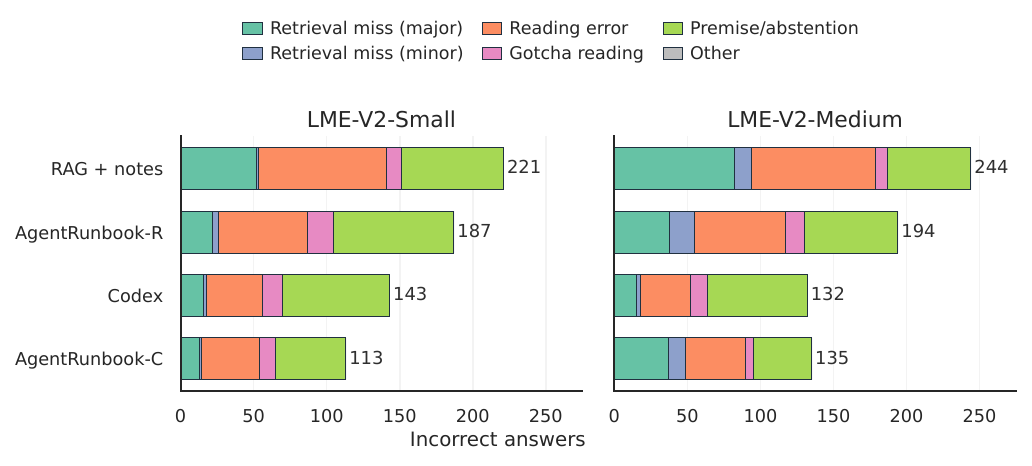}
\caption{Error composition for the main methods on LME-V2-Small and LME-V2-Medium. Each horizontal bar decomposes the incorrect answers for one method into retrieval misses, reading failures, gotcha failures, and premise-awareness failures.}
\label{fig:appendix-error-taxonomy}
\end{figure}

As shown in \Cref{fig:appendix-error-taxonomy}, AgentRunbook-R significantly reduces retrieval errors and downstream reading errors compared to the RAG+notes baseline. It does not improve abstention as it directly presents the relevant evidence to the downstream model without an analysis and thus the model can be misled into using the evidence to the question instead of rejecting it. Both Codex and AgentRunbook-C make fewer retrieval errors and downstream reader errors compared to the RAG methods. AgentRunbook-C further improves the abstention performance as the memory module is also instructed to explicitly identify the inconsistencies and wrong question premises and present them to the downstream model. 

\subsection{Tool Calling Behavior}
We compare AgentRunbook-C with the Codex baseline by parsing the coding agent event streams and grouping shell commands into five behavior classes. The \textit{setup/read task} group includes orientation and prompt/question reads. \textit{Harness-guided retrieval} includes manifest/summary reads and AgentRunbook helper calls. \textit{Raw trajectory exploration} includes broad
filesystem searches, direct trajectory reads, and ad-hoc Python scans. The remaining groups capture visual inspection and output validation or other commands.

\begin{figure}[t]
\centering
\includegraphics[width=0.9\linewidth]{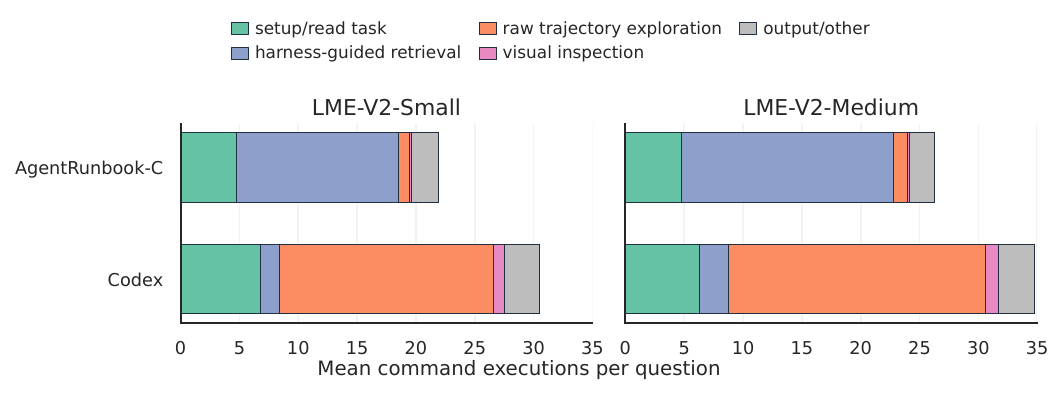}
\caption{Mean number of command executions per query divided by command types.}
\label{fig:appendix-command-counts}
\end{figure}

\begin{figure}[t]
\centering
\includegraphics[width=0.9\linewidth]{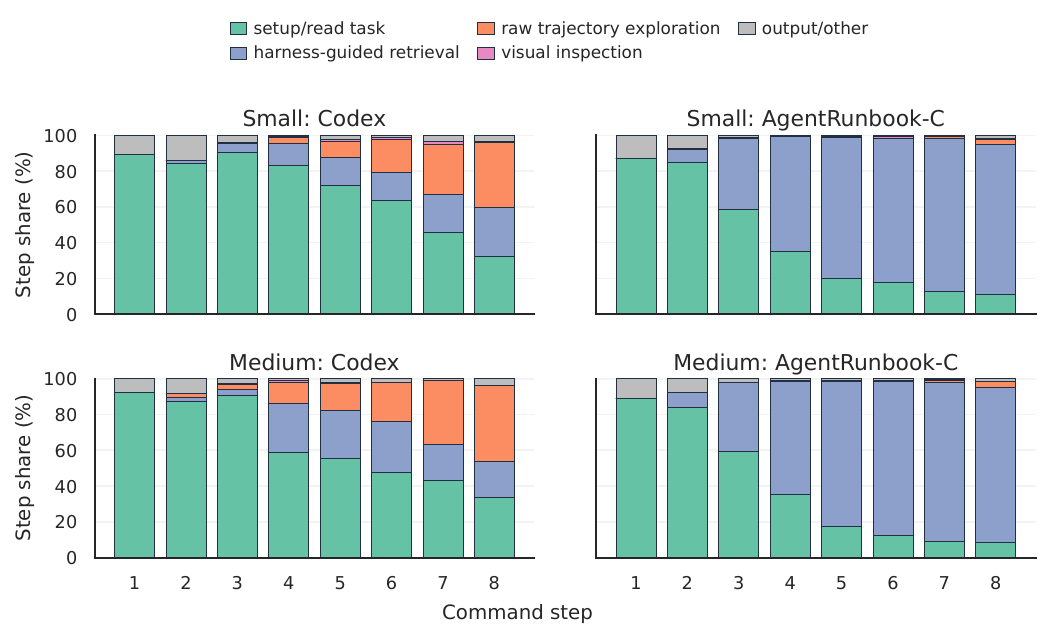}
\caption{Distribution of grouped command classes during the first eight command rounds.}
\label{fig:appendix-early-command-flow}
\end{figure}

As shown in \Cref{fig:appendix-command-counts}, AgentRunbook-C reduces the total number of commands and shifts work from raw trajectory exploration toward harness-guided retrieval. On LME-V2-Medium, Codex uses 21.8 raw trajectory-exploration commands per question on average, while AgentRunbook-C uses 18.0 harness-guided retrieval commands and only 1.2 raw trajectory exploration commands. In \Cref{fig:appendix-early-command-flow}, we further visualize the command distribution in the first rounds. Both methods begin with setup and task-reading commands. AgentRunbook-C then moves quickly into harness-guided retrieval, while Codex increasingly falls back to raw trajectory exploration in later early rounds.

\subsection{Qualitative Examples} 
Finally, we present qualitative examples for the two AgentRunbook variants from some successful queries in \Cref{tab:appendix-agentrunbook-c-span-examples} and \Cref{tab:appendix-agentrunbook-r-pool-examples}. For AgentRunbook-R, we select one example where each memory pool provides the answer-bearing evidence: procedure/hint notes, state-transition events, and raw state slices. For AgentRunbook-C, we show two selected evidence spans from different question types to illustrate the form of evidence passed to the
reader.

\newcolumntype{L}[1]{>{\raggedright\arraybackslash}p{#1}}
\newcolumntype{Y}{>{\raggedright\arraybackslash}X}
\newcolumntype{C}[1]{>{\centering\arraybackslash}p{#1}}

\begin{table*}[p]
\centering
\scriptsize
\setlength{\tabcolsep}{4pt}
\renewcommand{\arraystretch}{1.14}
\caption{Successful AgentRunbook-C examples with the selected question markdown, evidence span content, and corresponding screenshot.}
\label{tab:appendix-agentrunbook-c-span-examples}
\begin{tabularx}{\linewidth}{L{0.105\linewidth}L{0.275\linewidth}Y C{0.185\linewidth}}
\toprule
\textbf{Question type} &
\textbf{Question markdown} &
\textbf{Selected span content} &
\textbf{Screenshot} \\
\midrule
\vspace{0pt}Static environment &
\vspace{0pt}\textbf{QID: \texttt{98b62f3d}.} \textbf{Question.} I am using our reddit-based custom forum website. For the create submission form, what are the names of the mandatory fields? Mark your final answer as a comma-separated list of short phrases in \texttt{\textbackslash boxed\{\}}.

\vspace{2pt}
\textbf{Gold / reader answer.} \textit{Title, Forum}. &
\vspace{0pt}\textbf{Support analysis.} The clearest evidence is trajectory \texttt{4ba5e9cb}, state 2, on the \textit{Create submission} form for \texttt{/submit/pittsburgh}. That state marks \textit{Title} and \textit{Forum} as required, while \textit{Body} is optional and the URL/Image controls are submission-type selectors.

\vspace{2pt}
\textbf{Trajectory state span.} \texttt{4ba5e9cb}: states 2--2.

\vspace{2pt}
\textbf{State 2 AXTree excerpt.}
\texttt{[136]} LabelText: \textit{Title} \texttt{*};
\texttt{[138]} textbox \textit{Title This field is required.}, required;
\texttt{[141]} LabelText: \textit{Body};
\texttt{[143]} textbox \textit{Body};
\texttt{[148]} checkbox \textit{Formatting help +};
\texttt{[277]} LabelText: \textit{Forum} \texttt{*};
\texttt{[384]} combobox value \textit{pittsburgh}. &
\vspace{0pt}
\begin{minipage}[t]{\linewidth}
\centering
\includegraphics[width=\linewidth,height=1.05in,keepaspectratio]{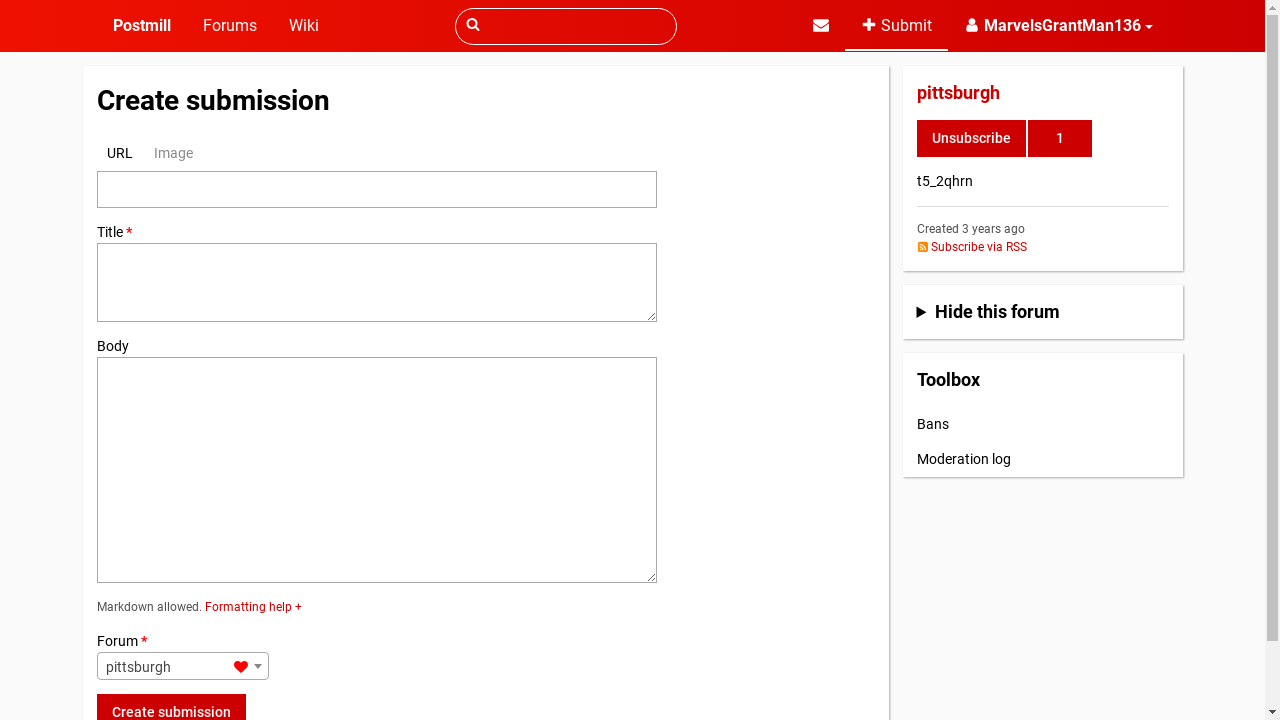}\\[-1pt]
\footnotesize Trajectory \texttt{4ba5e9cb}, state 2.
\end{minipage}
\\
\midrule
\vspace{0pt}Dynamic environment &
\vspace{0pt}\textbf{QID: \texttt{609acb91}.} \textbf{Question.} I am using our magento-based custom shopping website. I am now browsing the item list that appears after I click a specific item category from the home page. If I then narrow the display scope by selecting a specific price range in the left column, two new links will appear after the selected range. What are the names of the links? Your final answer should be a comma-separated list of two phrases wrapped in \texttt{\textbackslash boxed\{\}}.

\vspace{2pt}
\textbf{Gold / reader answer.} \textit{Remove This Item, Clear All}. &
\vspace{0pt}\textbf{Support analysis.} The supporting evidence is trajectory \texttt{dddd8aa2}, state 5. That state shows the Men $>$ Shoes category page after the \textit{Price: \$0.00--\$29.99} filter is applied, and immediately after the selected range the sidebar lists the two links \textit{Remove This Item} and \textit{Clear All}.

\vspace{2pt}
\textbf{Trajectory state span.} \texttt{dddd8aa2}: states 5--5. The action sequence opens Men $>$ Shoes, selects \textit{Price}, and loads the filtered page with \texttt{price=0-30}.

\vspace{2pt}
\textbf{State 5 AXTree excerpt.}
\texttt{[1922]} strong: \textit{Shop By};
\texttt{[1925]} heading: \textit{Now Shopping by};
\texttt{[1927]} listitem: \textit{Price: \$0.00--\$29.99};
\texttt{[1930]} link: \textit{Remove This Item}, clickable, visible;
\texttt{[1933]} link: \textit{Clear All}, clickable, visible. &
\vspace{0pt}
\begin{minipage}[t]{\linewidth}
\centering
\includegraphics[width=\linewidth,height=1.05in,keepaspectratio]{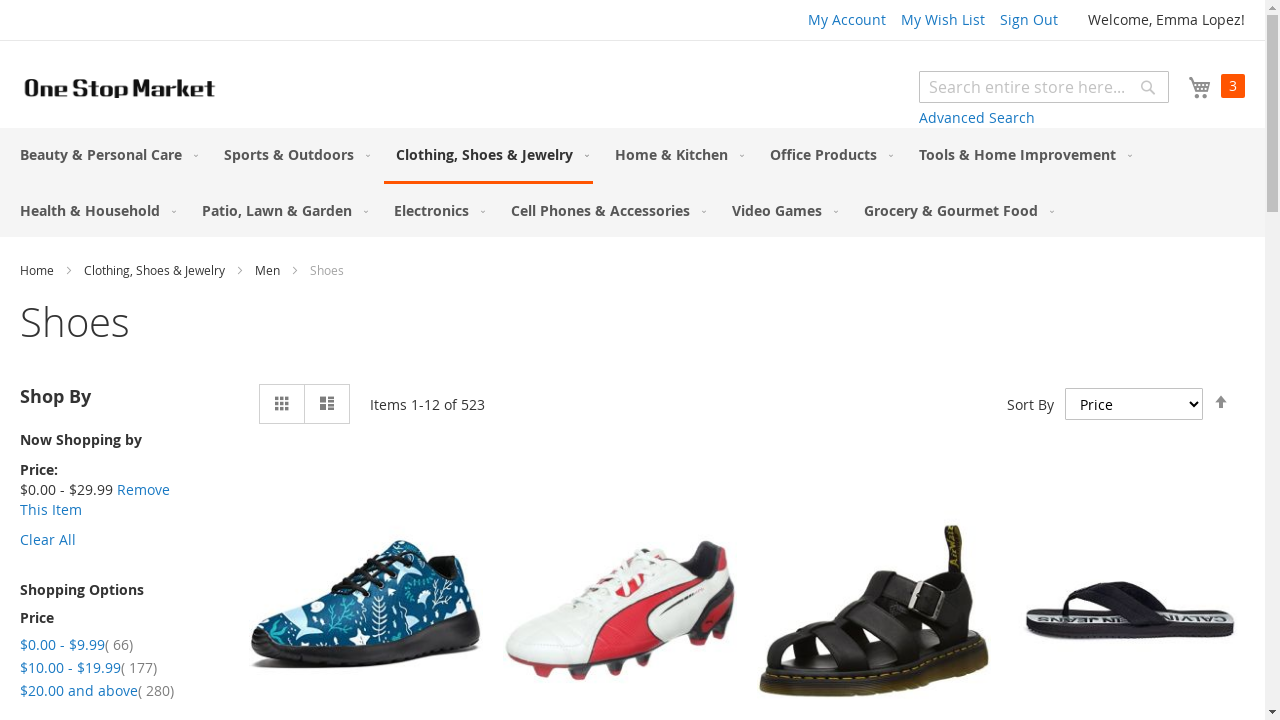}\\[-1pt]
\footnotesize Trajectory \texttt{dddd8aa2}, state 5.
\end{minipage}
\\
\bottomrule
\end{tabularx}
\end{table*}

\begin{table*}[p]
\centering
\scriptsize
\setlength{\tabcolsep}{3pt}
\renewcommand{\arraystretch}{1.14}
\caption{Successful AgentRunbook-R examples where different memory pools provide the answer-bearing evidence. The controller queries and retrieved items are copied from the corresponding runs.}
\label{tab:appendix-agentrunbook-r-pool-examples}
\begin{tabularx}{\linewidth}{L{0.105\linewidth}L{0.215\linewidth}L{0.170\linewidth}Y C{0.185\linewidth}}
\toprule
\textbf{Pool} &
\textbf{Question} &
\textbf{Controller query} &
\textbf{Retrieved evidence from that pool} &
\textbf{Figure} \\
\midrule

\vspace{0pt}Procedure and hint notes &
\vspace{0pt}Magento storefront order-history question: the user is already on \textit{My Orders}; the table says ``Items 1 to 10 of 37 total''; which pagination label should be clicked first to reach the oldest orders most directly? Choices: A. 2, B. 3, C. 4, D. Next, E. Last. \textbf{Correct answer: C.} &
\vspace{0pt}\textit{Magento My Orders page pagination and date sorting behavior for finding oldest purchase} &
\vspace{0pt}\textbf{Procedure note, rank 1} (score 0.7465, trajectory \texttt{19022110}) is titled \textit{Find Earliest Purchase Date in My Orders}. It states that on \textit{My Orders}, the reader should note the pagination text, compute the final page from the total count, click the final page number, and verify the final range, e.g., ``Items 31 to 37 of 37 total.'' The paired hint note also records that oldest orders are on the last page. With 37 orders and 10 per page, the first direct click is page 4. &
\vspace{0pt}\begin{minipage}[t]{\linewidth}
\centering
\includegraphics[width=\linewidth,height=1.05in,keepaspectratio]{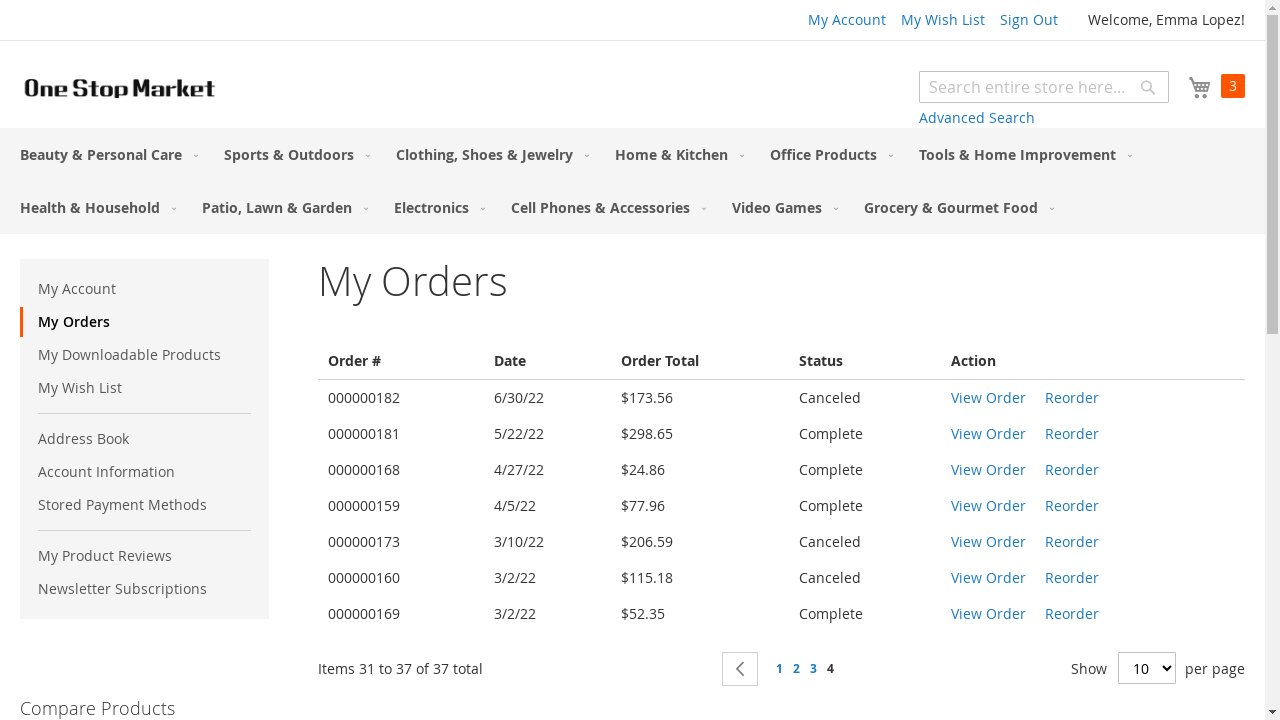}\\[-1pt]
\footnotesize Trajectory \texttt{19022110}: final order-history page.
\end{minipage}
\\
\midrule

\vspace{0pt}State-transition events &
\vspace{0pt}Postmill forum dynamic question: after replying to a nested comment, a blue banner appears above the reply; what does the link in that banner say? \textbf{Correct answer: View all comments.} &
\vspace{0pt}\textit{submit comment reply to nested thread and observe blue banner overlay appearance} &
\vspace{0pt}\textbf{Event result 2} (similarity 0.5923, trajectory \texttt{2e8f6477}) retrieves the transition from state 9 to state 10 after the nested reply was posted. The event is attached to the single-comment-thread view for the AskReddit post and includes the post-state screenshot where the banner text is visible. The banner reads ``Viewing a single comment thread.'' and its link reads ``View all comments.'' &
\vspace{0pt}\begin{minipage}[t]{\linewidth}
\centering
\includegraphics[width=\linewidth,height=1.05in,keepaspectratio]{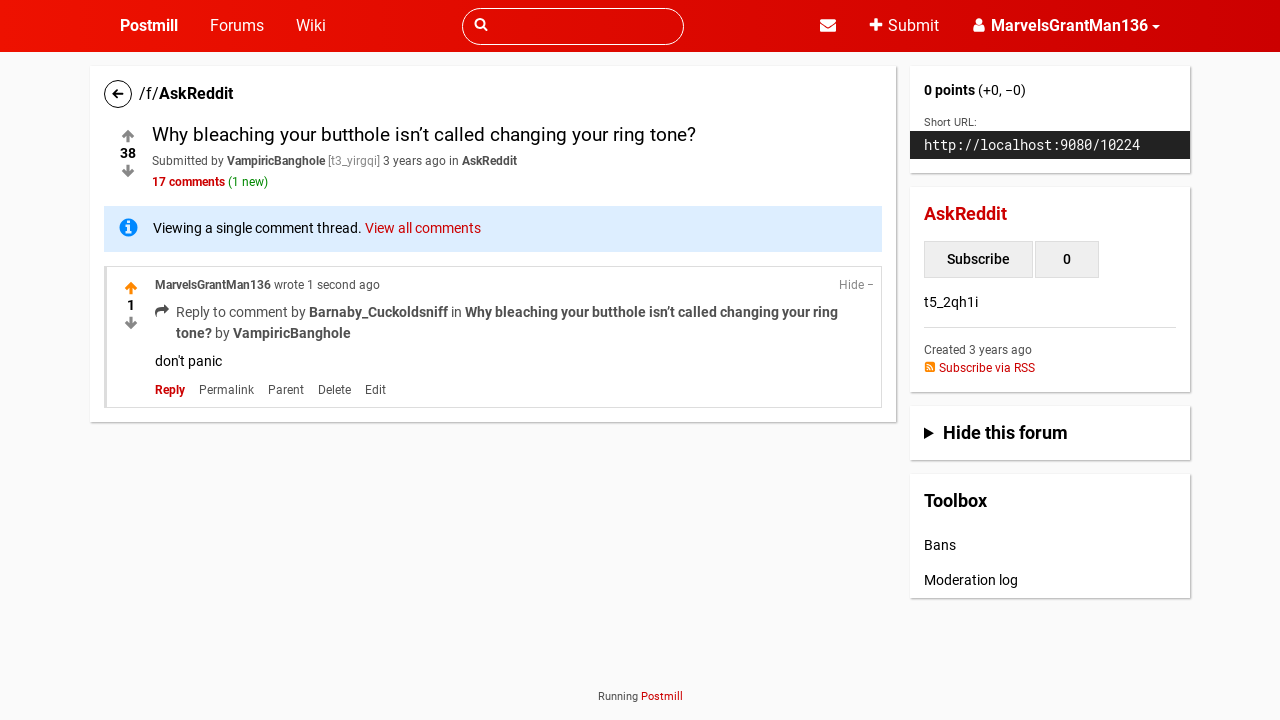}\\[-1pt]
\footnotesize Trajectory \texttt{2e8f6477}: reply result banner.
\end{minipage}
\\
\midrule

\vspace{0pt}Raw state slices &
\vspace{0pt}ServiceNow form-comparison question: between \textit{Create Change Request} and \textit{Incident}, what additional top-right button appears on the incident form but not the change-request form? \textbf{Correct answer: Resolve.} &
\vspace{0pt}\textit{ServiceNow Incident form top right button area visible controls}; \textit{ServiceNow Change Request form top right button area visible controls} &
\vspace{0pt}\textbf{Raw state result 1} (similarity 0.7168, trajectory \texttt{454485ca}, center state 7) retrieves the incident creation form; the top-right controls include \textit{Submit} and \textit{Resolve}. \textbf{Raw state result 4} (similarity 0.7087, trajectory \texttt{afa62eac}, center state 6) retrieves the change-request creation form; its top-right controls show \textit{Submit} without \textit{Resolve}. The contrast isolates the extra incident-only button. &
\vspace{0pt}\begin{minipage}[t]{\linewidth}
\centering
\includegraphics[width=\linewidth,height=0.78in,keepaspectratio]{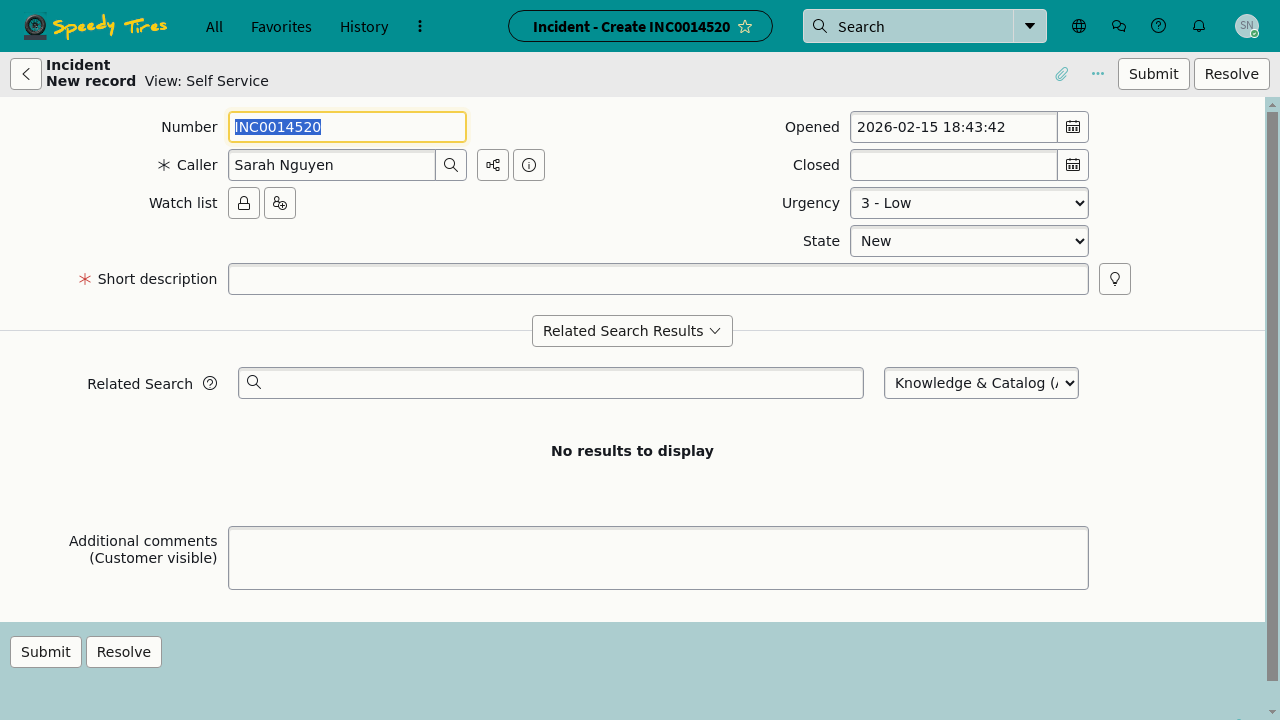}\\[-1pt]
\footnotesize Incident: \textit{Submit}, \textit{Resolve}.\\[2pt]
\includegraphics[width=\linewidth,height=0.78in,keepaspectratio]{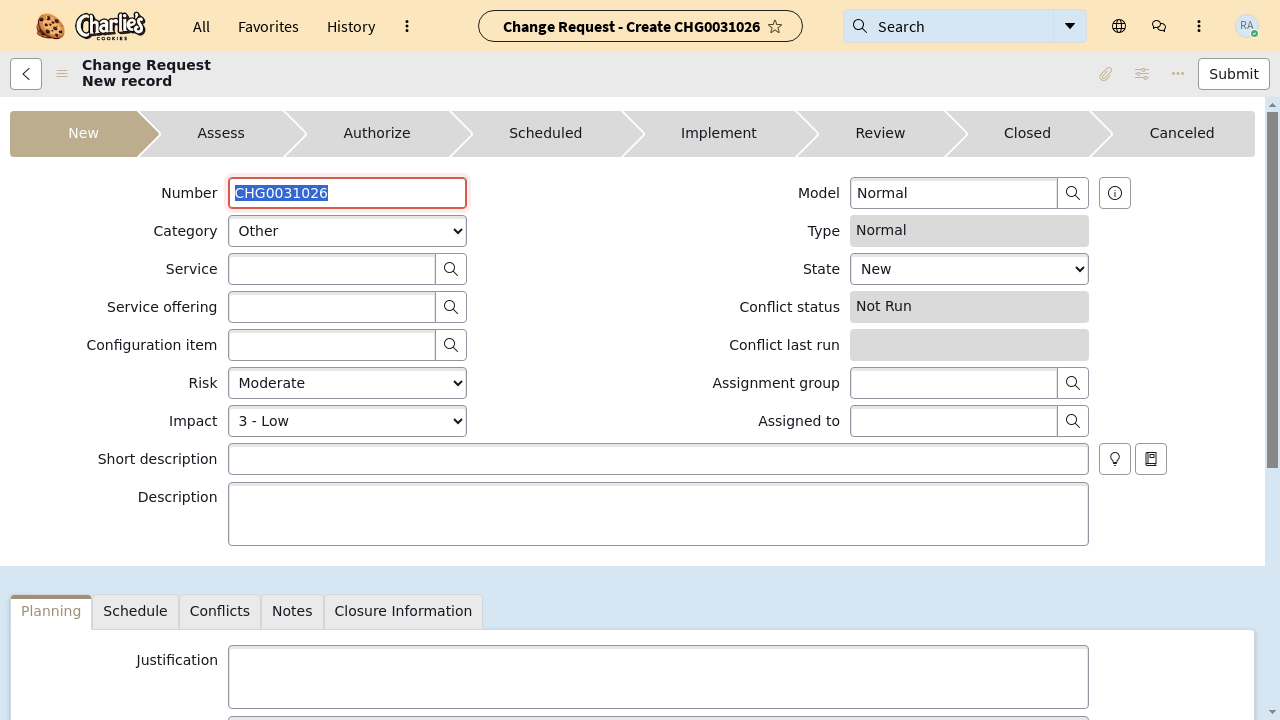}\\[-1pt]
\footnotesize Change Request: \textit{Submit} only.
\end{minipage}
\\

\bottomrule
\end{tabularx}
\end{table*}

\section{Limitations and Ethics Statements}

\subsection{Limitations}
\label{section-limitations}

\paragraph{Benchmark scope and formulation.}
LME-V2 focuses on web agents operating in customized browser environments. Browser use is a broad and practically important domain, but it does not cover the full space of digital agents, such as coding agents, computer-use agents, or domain-specific enterprise agents, which may involve different memory requirements and risk profiles. In addition, for reproducibility and controlled comparison, LME-V2 evaluates memory over pre-collected trajectory histories rather than online learning during live task execution. This design makes the benchmark easier to distribute and reproduce, but it may not fully capture distribution shifts caused by an agent's own evolving behavior. Finally, our context gathering formulation evaluates whether a memory module can return useful evidence for a fixed reader model, rather than directly measuring end-to-end task success. This is an intentional design choice to isolate memory quality, but downstream agent correctness may also depend on planning, tool use, and action execution.

\paragraph{Methodological scope.}
AgentRunbook explores practical memory designs built around retrieval, file organization, and agent-native evidence inspection, rather than proposing new model architectures or training procedures. While this keeps the methods simple and easy to analyze, their performance depends on the quality of trajectory representations, retrieval models, prompts, and coding agent behavior. AgentRunbook-C significantly improves over an off-the-shelf Codex harness in our experiments, but we do not build a fully customized coding agent harness from scratch. Future work could study tighter integration between memory, planning, and execution, as well as learned memory controllers that adapt their storage and retrieval strategies across environments.

\subsection{Ethics Statements}
\label{section-ethics-statements}
\label{section-broader-impacts}

\paragraph{Human annotators.}
The benchmark was constructed by four student authors, consisting of one graduate student and three undergraduate students. The graduate student was compensated through wage support, and the undergraduate students received research credits; all annotators are also credited through authorship. The annotation process included regular weekly annotation sessions, question-review discussions, and trajectory-label verification meetings with the project lead. Annotators were briefed on the research goals and downstream use of their annotations, and they provided consent for their annotation work to be used in the benchmark. Under the institution's research policy, this work was determined to be exempt from IRB approval.

\paragraph{Dataset privacy, bias, and safeguards for reuse.}
LME-V2 is built from sandboxed web-agent environments derived from WebArena, WorkArena, and WorkArena++, rather than from real user browsing histories. The environments use synthetic tasks, names, records, and personal information, so the risk of exposing real private user data is minimal. We manually inspected the curated LME-V2 questions and did not identify additional information leakage beyond the intended synthetic environment content. We also rely on the upstream benchmark creators' sanitization of their released environments and assets. Since the benchmark primarily covers synthetic professional workflows and controlled web environments, it is not intended to represent real demographic populations and is unlikely to directly propagate or amplify demographic bias. Nevertheless, released artifacts will be documented with intended-use guidance, and users should not treat LME-V2 as evidence that memory systems are safe for deployment on real user histories or sensitive enterprise data without additional privacy review, access control, and data minimization safeguards.

\paragraph{Dataset and artifact licensing.}
We respect the intended use and license terms of the upstream artifacts used in this work. WebArena, AgentLab, and the Codex GitHub repository are released under the Apache-2.0 license, and our use of these assets follows their respective licensing requirements. For Codex-based experiments, we use OpenAI API access rather than commercial ChatGPT subscriptions to avoid violating consumer product usage terms. We plan to release our code and derived benchmark artifacts under the Apache-2.0 license as well. We do not redistribute the original benchmark datasets or the agent harnesses used for trajectory collection. Instead, we release the derived trajectory traces, including accessibility trees and screenshots, as necessary benchmark artifacts for reproducing and evaluating LME-V2.

\paragraph{Broader societal impacts.}
This work aims to improve the efficiency and reliability of long-running agents by evaluating and designing memory systems that reuse prior experience instead of repeatedly rediscovering the same environment knowledge. More effective memory may reduce redundant agent computation, which can lower cost and environmental impact, and may also support self-improving systems that develop useful expertise in specialized domains. At the same time, persistent memory introduces risks. Agents may drift in behavior if they rely on stale, incorrect, or context-dependent observations, and coding agent-based memory controllers require careful sandboxing because their file system and tool use abilities can introduce additional security risks. More capable self-learning agents may also accelerate automation in economically valuable workflows, which could contribute to labor displacement. These risks reinforce the need to evaluate memory systems under controlled settings, document intended uses, and apply deployment safeguards such as sandboxing, access control, memory expiration, and human oversight.

\subsection{LLM and Agent Use}
\label{section-llm-use}

No LLMs or agents were used for research ideation. The benchmark questions in LME-V2 were written by human annotators. As described in the main paper and appendix, we used Codex to generate initial proposals for the large-scale answer trajectory labeling which were then verified by humans. We also used Codex to assist with experiment implementation and all generated code was reviewed and validated by the authors. Codex was used to generate the figures and tables in the paper from human-provided data, except for \Cref{fig:main-examples} and \Cref{fig:agentrunbook-methods}. ChatGPT is used to generate the illustration in \Cref{fig:agentrunbook-methods}, which we rigorously validate to ensure its consistency with the actual method implementation. ChatGPT, specifically GPT-5.5, is used to polish the paper from a purely human-written draft. The authors take full responsibility for the content, claims, experiments, code, figures, and tables in this submission.


\end{document}